\documentclass[10pt,twocolumn,letterpaper]{article}
\pdfoutput=1

\usepackage{cvpr}
\usepackage{times}
\usepackage{epsfig}
\usepackage{graphicx}
\usepackage{amsmath}
\usepackage{amssymb}
\usepackage{mathtools}
\usepackage[normalem]{ulem}

\usepackage{pifont}

\usepackage[pagebackref=true,breaklinks=true,letterpaper=true,colorlinks,bookmarks=false]{hyperref}

\usepackage[usenames,dvipsnames]{xcolor}
\definecolor{dgreen}{rgb}{0.0,0.6,0.0} 
\definecolor{dred}{rgb}{0.6,0.0,0.0} 
\definecolor{alexey}{rgb}{0.7,0,1}
\definecolor{philipcolor}{rgb}{0,0.5,0}
\definecolor{grey}{rgb}{0.6,0.6,0.6}

\newcommand{\my}{\textcolor{dgreen}{\ding{51}}}%
\newcommand{\mn}{\textcolor{dred}{\ding{55}}}%
\newcommand{\msp}{\small{sparse}}%

\graphicspath{{./images/}}

\usepackage[pagebackref=true,breaklinks=true,letterpaper=true,colorlinks,bookmarks=false]{hyperref}
\hypersetup{citecolor=red} 
\hypersetup{linkcolor=dgreen} 
\hypersetup{pdftitle={A Large Dataset to Train Convolutional Networks for Disparity, Optical Flow, and Scene Flow Estimation},
            pdfauthor={Nikolaus Mayer, Eddy Ilg, Philip H{\"a}usser, Philipp Fischer, Daniel Cremers, Alexey Dosovitskiy, Thomas Brox},
            pdfsubject={CVPR 2016 Submission and Supplementary Material (arXiv.org version)},
            pdfkeywords={disparity, optical flow, scene flow, synthetic, datasets, convolutional neural networks, cnn}}

\cvprfinalcopy 

\def\cvprPaperID{280} 

\ifcvprfinal\fi
\begin{document}

\title{A Large Dataset to Train Convolutional Networks\\[1mm] for Disparity, Optical Flow, and Scene Flow Estimation}

\renewcommand*{\thefootnote}{\fnsymbol{footnote}}

\author{
  Nikolaus Mayer$^*$$^1$,
  Eddy Ilg$^*$$^1$,
  Philip H{\"a}usser$^*$$^2$,
  Philipp Fischer$^*$$^1$\stepcounter{footnote}\footnote{Supported by the Deutsche Telekom Stiftung}\\
  $^1$University of Freiburg \hspace*{1cm} $^2$Technical University of Munich\\
  {\tt\small $^1$\{mayern,ilg,fischer\}@cs.uni-freiburg.de \hspace{.5cm} $^2$haeusser@cs.tum.edu}
\and
  Daniel Cremers\\
  Technical University of Munich\\
  {\tt\small cremers@tum.de}
\and
  Alexey Dosovitskiy, 
  Thomas Brox\\
  University of Freiburg\\
  {\tt\small \{dosovits,brox\}@cs.uni-freiburg.de}
}


\maketitle

\stepcounter{footnote}\footnotetext{These authors contributed equally}
\stepcounter{footnote}\footnotetext{Supported by the Deutsche Telekom Stiftung}

\renewcommand*{\thefootnote}{\arabic{footnote}}

\begin{abstract}
Recent work has shown that optical flow estimation can be formulated as a supervised learning task and can be successfully solved with convolutional networks. 
Training of the so-called FlowNet was enabled by a large synthetically generated dataset. The present paper extends the concept of optical flow estimation via convolutional networks to disparity and scene flow estimation. 
To this end, we propose three synthetic stereo video datasets with sufficient realism, variation, and size to successfully train large networks. 
Our datasets are the first large-scale datasets to enable training and evaluating scene flow methods. 
Besides the datasets, we present a convolutional network for real-time disparity estimation that provides state-of-the-art results. By combining a flow and disparity estimation network and training it jointly, we demonstrate the first scene flow estimation with a convolutional network. 
\end{abstract}

\section{Introduction}\label{sec:intro}
Estimating scene flow means providing the depth and 3D motion vectors of all visible points in a stereo video. 
It is the ``royal league'' task when it comes to reconstruction and motion estimation and provides an important basis for numerous higher-level challenges such as advanced driver assistance and autonomous systems.
Research over the last decades has focused on its subtasks, namely disparity estimation and optical flow estimation, with considerable success.  
The full scene flow problem has not been explored to the same extent. 
While partial scene flow can be simply assembled from the subtask results, it is expected that the joint estimation of all components would be advantageous, with regard to both efficiency and accuracy.
One reason for scene flow being less explored than its subtasks seems to be a shortage of fully annotated ground truth data.

The availability of such data has become even more important in the era of convolutional networks. 
Dosovitskiy et~al.~\cite{FlowNet} showed that optical flow estimation can be posed as a supervised learning problem and can be solved with a large network. 
For training their network, they created a simple synthetic 2D dataset of flying chairs, which proved to be sufficient to predict accurate optical flow in general videos. 
These results suggest that also disparities and scene flow can be estimated via a convolutional network, ideally jointly, efficiently, and in real-time.
What is missing to implement this idea is a large dataset with sufficient realism and variability to train such a network and to evaluate its performance. 


\begin{figure}[t]
  \begin{center}
  {
    \setlength{\tabcolsep}{1pt}%
    \begin{tabular}{ccc}
      \multicolumn{3}{c}{ \includegraphics[width=0.92\linewidth]{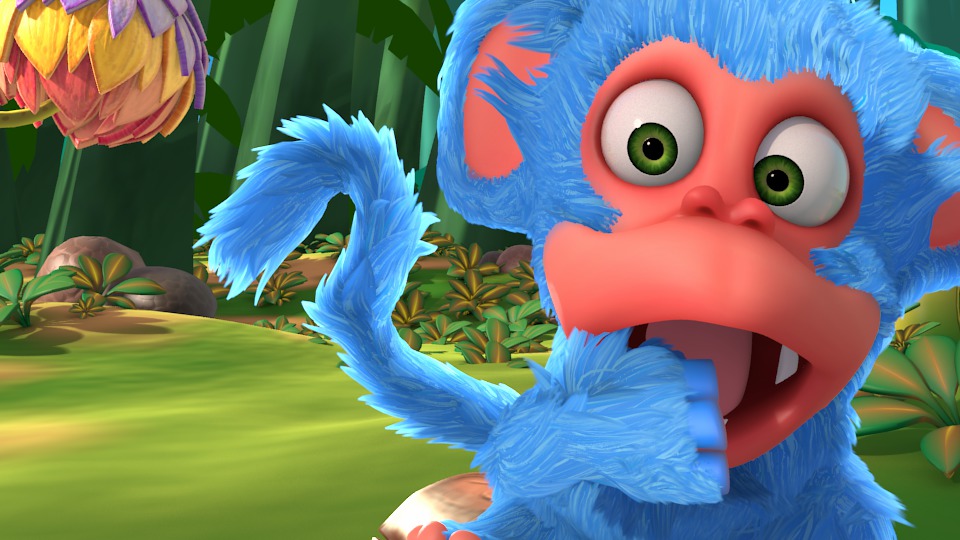} } \\
      \includegraphics[width=0.3\linewidth]{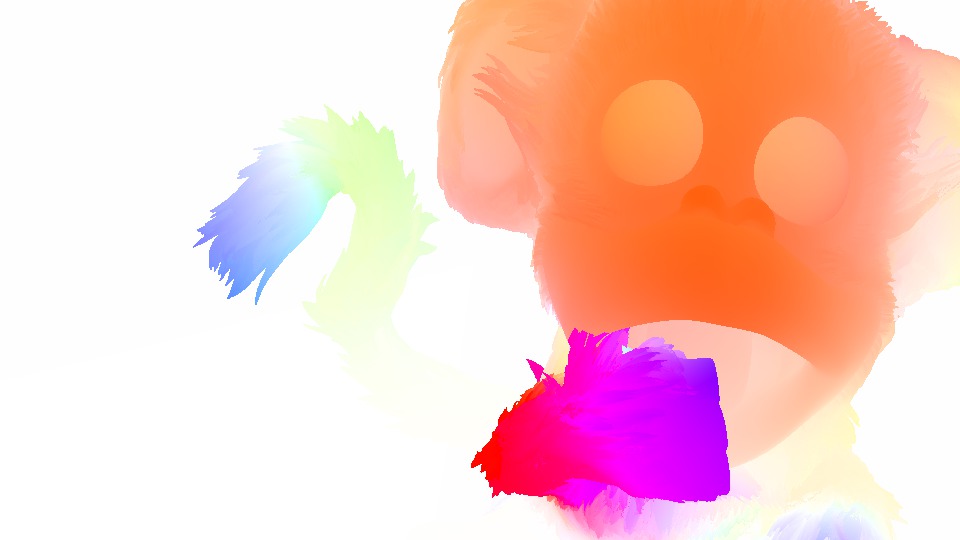} %
      & \includegraphics[width=0.3\linewidth]{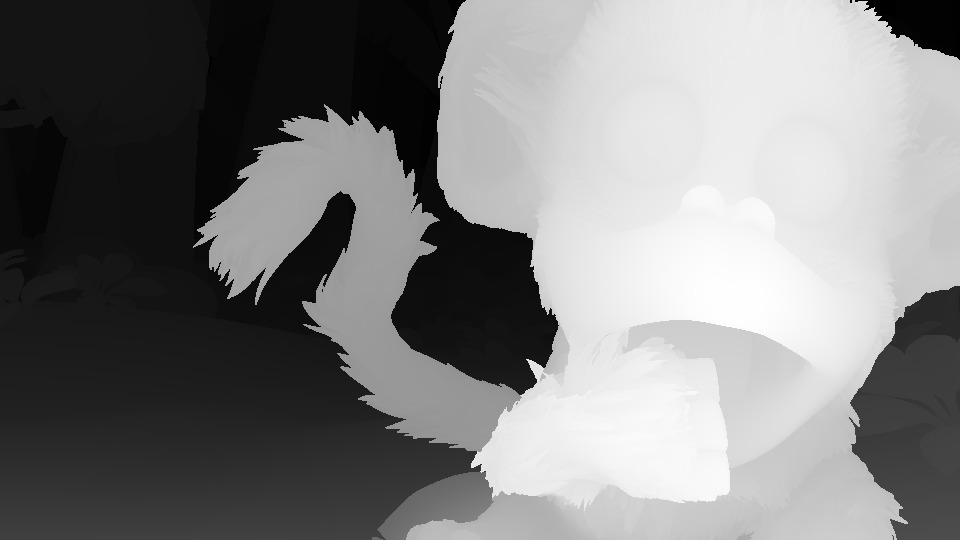} %
      & \includegraphics[width=0.3\linewidth]{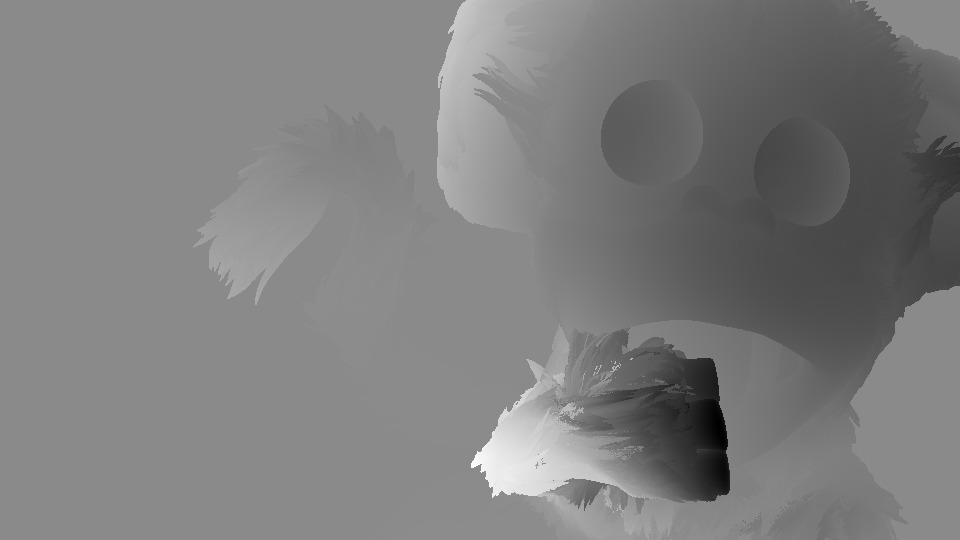} \\
    \end{tabular}
  }
  \end{center}
  \caption{Our datasets provide over $35\,000$ stereo frames with dense ground truth for \emph{optical flow}, \emph{disparity} and \emph{disparity change}, as well as other data such as object segmentation.}
  \label{fig:teaser}
\end{figure}

\begin{table*}[t]
  \begin{center}
  \resizebox{0.98\textwidth}{!}{%
  \begin{tabular}{|l||c|cc|c|c|ccc|}
      \hline
      Dataset & MPI Sintel \cite{Butler-et-al-12} & \multicolumn{2}{c|}{KITTI Benchmark Suite \cite{Menze2015CVPR}} & SUN3D\cite{sun3d} & NYU2\cite{nyu2} & \multicolumn{3}{c|}{Ours}\\
                         &                      & 2012                 & 2015   &        &         & FlyingThings3D & Monkaa & Driving\\
      \hline\hline
      \#Training frames  & $1\,064$             & $194$                & $800$  & $2.5$M & $1\,449$ & $21\,818$ & $8\,591$ & $4\,392$ \\
      \#Test frames      & $564$                & $195$                & $800$  & ---    & ---     &  $4\,248$ & ---     & --- \\
      \#Training scenes  & $25$                 & $194$                & $200$  & $415$  & $464$   &  $2\,247$ & $8$     & $1$ \\
      Resolution         & $1024\!\times\!436$  & $1226\!\times\!370$  & $1242\!\times\!375$ & $640\!\times\!480$ & $640\!\times\!480$  & $960\!\times\!540$ & $960\!\times\!540$   & $960\!\times\!540$ \\
      \hline
      Disparity/Depth    & \my   & \msp   & \msp   & \my    & \my   & \my   & \my   & \my   \\
      Disparity change   & \mn   & \mn    & \mn    & \mn    & \mn   & \my   & \my   & \my   \\
      Optical flow       & \my   & (\msp) & (\msp) & \mn    & \mn   & \my   & \my   & \my   \\
      \hline  
      Segmentation       & \my   & \mn    & \mn    & (\my)  & \my   & \my   & \my   & \my   \\
      Motion boundaries  & \my   & \mn    & \mn    & \mn    & \mn   & \my   & \my   & \my   \\
      Naturalism         & (\my) & \my    & \my    & \my    & \my   & \mn   & \mn   & (\my) \\
      \hline
    \end{tabular}}
  \end{center}
  \caption{Comparison of available datasets: Our new collection offers more annotated data and greater data variety than any existing choice. All our data has fully contiguous, dense, accurate ground truth.}
  \label{table:datasets_comparison}
\end{table*}

\pagebreak

In this paper, we present a collection of three such datasets, made using a customized version of the open source 3D creation suite Blender\footnote{\url{https://www.blender.org/}}. Our effort is similar in spirit to the Sintel benchmark \cite{Butler-et-al-12}. In contrast to Sintel, our dataset is large enough to facilitate training of convolutional networks, and it provides ground truth for scene flow. 
In particular, it includes stereo color images and ground truth for bidirectional disparity, bidirectional optical flow and disparity change, motion boundaries, and object segmentation.
Moreover, the full camera calibration and 3D point positions are available, i.e. our dataset also covers RGBD data. 

We cannot exploit the full potential of this dataset in a single paper, but we already demonstrate various usage examples in conjunction with convolutional network training. 
We train a network for disparity estimation, which yields competitive performance also on previous benchmarks, especially among those methods that run in real-time. 
Finally, we also present a network for scene flow estimation and provide the first quantitative numbers on full scene flow on a sufficiently sized test set.

\section{Related Work}\label{sec:related}

\textbf{Datasets.}
The first significant efforts to create standard datasets were the Middlebury datasets for stereo disparity estimation~\cite{scharstein2002taxonomy} and optical flow estimation~\cite{Baker-et-al-09}. While the stereo dataset consists of real scenes, the optical flow dataset is a mixture of real scenes and rendered scenes. Both datasets are very small in today's terms. Especially the small test sets have led to heavy manual overfitting. An advantage of the stereo dataset is the availability of relevant real scenes, especially in the latest high-resolution version from 2014~\cite{scharstein2014high}.

MPI Sintel~\cite{Butler-et-al-12} is an entirely synthetic dataset derived from a short open source animated 3D movie. It provides dense ground truth for optical flow. Since very recently, a beta testing version of disparities is available for training. With $1064$ training frames, the Sintel dataset is the largest dataset currently available. It contains sufficiently realistic scenes including natural image degradations such as fog and motion blur. The authors put much effort into the correctness of the ground truth for all frames and pixels. This makes the dataset a very reliable test set for comparison of methods. However, for training convolutional networks, the dataset is still too small. 

The KITTI dataset was produced in 2012~\cite{Geiger-et-al-13} and extended in 2015~\cite{Menze2015CVPR}. It contains stereo videos of road scenes from a calibrated pair of cameras mounted on a car. Ground truth for optical flow and disparity is obtained from a 3D laser scanner combined with the egomotion data of the car. While the dataset contains real data, the acquisition method restricts the ground truth to static parts of the scene. Moreover, the laser only provides sparse data up to a certain distance and height. For the most recent version, 3D models of cars were fitted to the point clouds to obtain a denser labeling and to include also moving objects. However, the ground truth in these areas is still an approximation. 

Dosovitskiy et al.~\cite{FlowNet} trained convolutional networks for optical flow estimation on a synthetic dataset of moving 2D chair images superimposed on natural background images.
This dataset is large but limited to single-view optical flow. It does not contain 3D motions and is not yet publicly available.

Both the latest Sintel dataset and the KITTI dataset can be used to estimate scene flow with some restrictions. In occluded areas (visible in one frame but not in the other), ground truth for scene flow is not available. On KITTI, the most interesting component of scene flow, namely the 3D motion of points, is missing or approximated via fitted CAD models of cars. A comprehensive overview of the most important comparable datasets and their features is given in Table~\ref{table:datasets_comparison}.


\textbf{Convolutional networks.}
Convolutional networks~\cite{lecun1989backpropagation} have proven very successful for a variety of recognition tasks, such as image classification~\cite{Krizhevsky-et-al-12}. Recent applications of convolutional networks include also depth estimation from single images~\cite{Eigen-et-al-14}, stereo matching~\cite{zbontar2015stereo}, and optical flow estimation~\cite{FlowNet}. 

The FlowNet of Dosovitskiy et al.~\cite{FlowNet} is most related to our work. It uses an encoder-decoder architecture with additional crosslinks between contracting and expanding network parts, where the encoder computes abstract features from receptive fields of increasing size, and the decoder reestablishes the original resolution via an expanding up-convolutional architecture~\cite{Dosovitskiy-15}. 
We adapt this approach for disparity estimation. 

The disparity estimation method in {\v{Z}}bontar et al.~\cite{zbontar2015stereo} uses a Siamese network for computing matching distances between image patches. To actually estimate the disparity, the authors then perform cross-based cost aggregation~\cite{ZhangLL09} and semi-global matching (SGM)~\cite{Hirschmueller2008PAMI}. In contrast to our work, {\v{Z}}bontar et al. have no end-to-end training of a convolutional network on the disparity estimation task, with corresponding consequences for computational efficiency and elegance. 


\textbf{Scene flow.}
While there are hundreds of papers on disparity estimation and optical flow estimation, there are only a few on scene flow. None of them uses a learning approach. 

Scene flow estimation was popularized for the first time by the work of Vedula et al.~\cite{VBRCK05} who analyzed different possible problem settings. Later works were dominated by variational methods. Huguet and Devernay~\cite{HD07} formulated scene flow estimation in a joint variational approach. Wedel et al.~\cite{wedel2008efficient} followed the variational framework but decoupled the disparity estimation for larger efficiency and accuracy. Vogel et al.~\cite{Vogel2013ICCV} combined the task of scene flow estimation with superpixel segmentation using a piecewise rigid model for regularization.
Quiroga et al.~\cite{quiroga2012scene} extended the regularizer further to a smooth field of rigid motion. Like Wedel et al.~\cite{wedel2008efficient} they decoupled the disparity estimation and replaced it by the depth values of RGBD videos. 

The fastest method in KITTI's scene flow top 7 is from Cech et al.~\cite{Cech11} with a runtime of 2.4 seconds. 
The method employs a seed growing algorithm for simultaneous disparity and optical flow estimation.

\section{Definition of Scene Flow}\label{sec:def_sf}

Optical flow is a projection of the world's 3D motion onto the image plane.
Commonly, \emph{scene flow} is considered as the underlying 3D motion field that can be computed from stereo videos or RGBD videos. Assume two successive time frames $t$ and $t\!+\!1$ of a stereo pair, yielding four images $(I^{t}_L$, $I^{t}_R$, $I^{t+1}_L$, $I^{t+1}_R)$. Scene flow provides for each visible point in one of these four images the point's 3D position and its 3D motion vector \cite{VedulaSF99}.

\begin{figure}[t]
\begin{center}
   \includegraphics[width=0.8\linewidth]{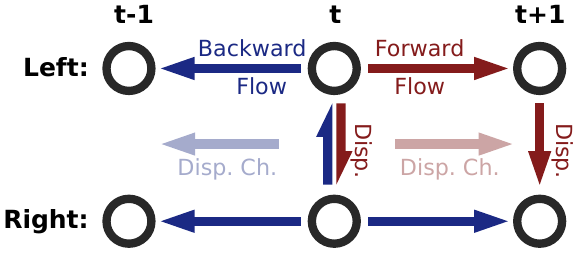}
\end{center}
   \caption{Given stereo images at times $t\!-\!1$, $t$ and $t\!+\!1$, the arrows indicate disparity and flow relations between them. The red components are commonly used to estimate scene flow. In our datasets we provide all relations including the blue arrows.}
\label{fig:the-h}
\end{figure}

These 3D quantities can be computed only in the case of known camera intrinsics and extrinsics.
A camera-independent definition of scene flow is obtained by the separate components \emph{optical flow}, the \emph{disparity}, and the \emph{disparity change}~\cite{HD07}, cf. Fig.~\ref{fig:the-h}.
This representation is complete in the sense that the visible 3D points and their 3D motion vectors can be computed from the components if the camera parameters are known. 

Given the disparities at $t$ and $t\!+\!1$, the disparity change is almost redundant.
Thus, in the KITTI 2015 scene flow benchmark \cite{Menze2015CVPR}, only optical flow and disparities are evaluated.
In this case, scene flow can be reconstructed only for surface points that are visible in both the left and the right frame. 
Especially in the context of convolutional networks, it is particularly interesting to estimate also depth and motion in partially occluded areas. 
Moreover, reconstruction of the 3D motions from flow and disparities is more sensitive to noise, because a small error in the optical flow can lead to a large error in the 3D motion vector.






\section{Three Rendered Datasets}\label{sec:datasets}

\label{sec:our_datasets}

We created a synthetic dataset suite that consists of three subsets and provides the complete ground truth scene flow (incl. disparity change) in forward and backward direction.  
To this end, we used the open source 3D creation suite Blender to animate a large number of objects with complex motions and to render the results into tens of thousands of frames.
We modified the pipeline of Blender's internal render engine to produce -- besides stereo RGB images -- three additional data passes per frame and view. 
These provide 3D positions of all visible surface points, as well as their future and past 3D positions. The pixelwise difference between two such data passes for a given camera view results in an "image" of 3D motion vectors -- the complete scene flow ground truth as seen by this camera. 
Note that the information is complete even in occluded regions since the render engine always has full knowledge about all (visible and invisible) scene points.
All non-opaque materials -- notably, most car windows -- were rendered as fully transparent to avoid consistency problems in the 3D data.

Given the intrinsic camera parameters (focal length, principal point) and the render settings (image size, virtual sensor size and format), we project the 3D motion vector of each pixel into a 2D pixel motion vector coplanar to the imaging plane: 
the optical flow. 
Depth is directly retrieved from a pixel's 3D position and converted to disparity using the known configuration of the virtual stereo rig.
We compute the \emph{disparity change} from the depth component of the 3D motion vector.
An example of the results is shown in Fig.~\ref{fig:teaser}.

In addition, we rendered object segmentation masks in which each pixel's value corresponds to the unique index of its object. 
Objects can consist of multiple subparts, of which each can have a separate \emph{material} (with own appearance properties such as textures). We make use of this and render additional segmentation masks, where each pixel encodes its material's index.
The recently available beta version of Sintel also includes this data.


Similar to the Sintel dataset, we also provide \emph{motion boundaries} which highlight pixels between at least two moving objects, if the following holds: 
The difference in motion between the two frames is at least $1.5$ pixels, and the boundary segment covers an area of at least $10$ pixels.
The thresholds were chosen to match the results of Sintel's segmentation.

For all frames and views, we provide the full camera intrinsics and extrinsics matrices. 
Those can be used for structure from motion or other tasks that require camera tracking.
We rendered all image data using a virtual focal length of $35$mm on a $32$mm wide simulated sensor.
For the \emph{Driving} dataset we added a wide-angle version using a focal length of $15$mm which is visually closer to the existing KITTI datasets.

Like the Sintel dataset, our datasets also include two distinct versions of every image: the \emph{clean pass} shows colors, textures and scene lighting but no image degradations, while the \emph{final pass} additionally includes postprocessing effects such as simulated depth-of-field blur, motion blur, sunlight glare, and gamma curve manipulation.

To handle the massive amount of data (2.5 TB), we compressed all RGB image data to the lossy but high-quality  WebP\footnote{\url{https://developers.google.com/speed/webp/}} format.
Non-RGB data was compressed losslessly using LZO\footnote{\url{http://www.oberhumer.com/opensource/lzo/}}.

\subsection{FlyingThings3D}\label{sec:flyingthings3d}

The main part of the new data collection consists of everyday objects flying along randomized 3D trajectories.
We generated about $25\,000$ stereo frames with ground truth data.
Instead of focusing on a particular task (like KITTI) or enforcing strict naturalism (like Sintel), we rely on randomness and a large pool of rendering assets to generate orders of magnitude more data than any existing option, without running a risk of repetition or saturation.
Data generation is fast, fully automatic, and yields dense accurate ground truth for the complete scene flow task.
The motivation for creating this dataset is to facilitate training of large convolutional networks, which should benefit from the large variety.

\begin{figure}[t]
  \begin{center}
  {
    \setlength{\tabcolsep}{1pt}%
    \begin{tabular}{cccc}
      \multicolumn{2}{c}{ \includegraphics[width=0.49\linewidth]{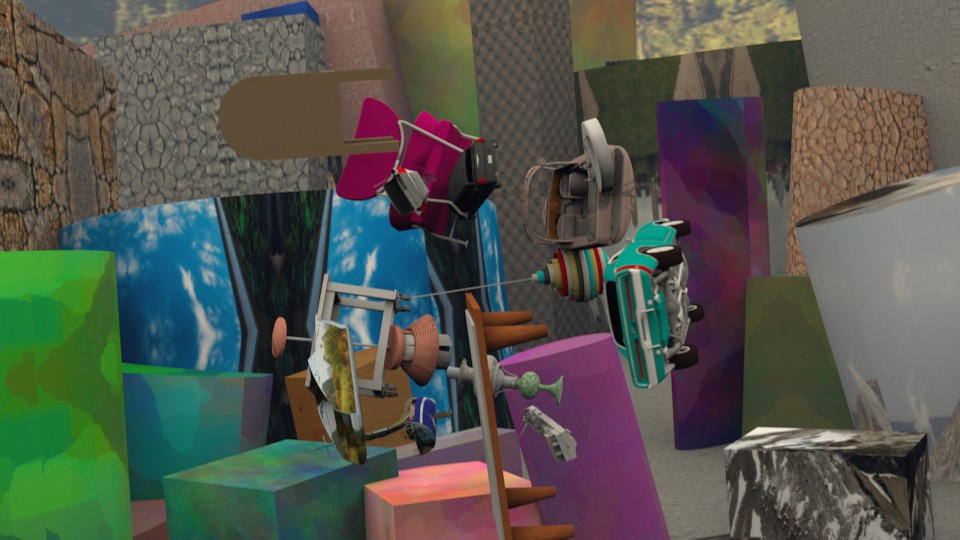} } &
      \multicolumn{2}{c}{ \includegraphics[width=0.49\linewidth]{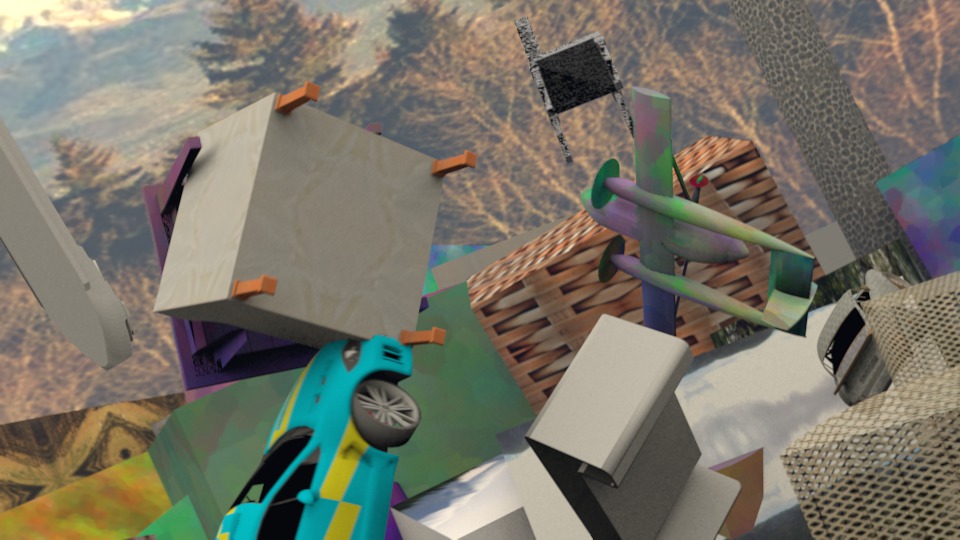} } \\
      
      \multicolumn{2}{c}{ \includegraphics[width=0.49\linewidth]{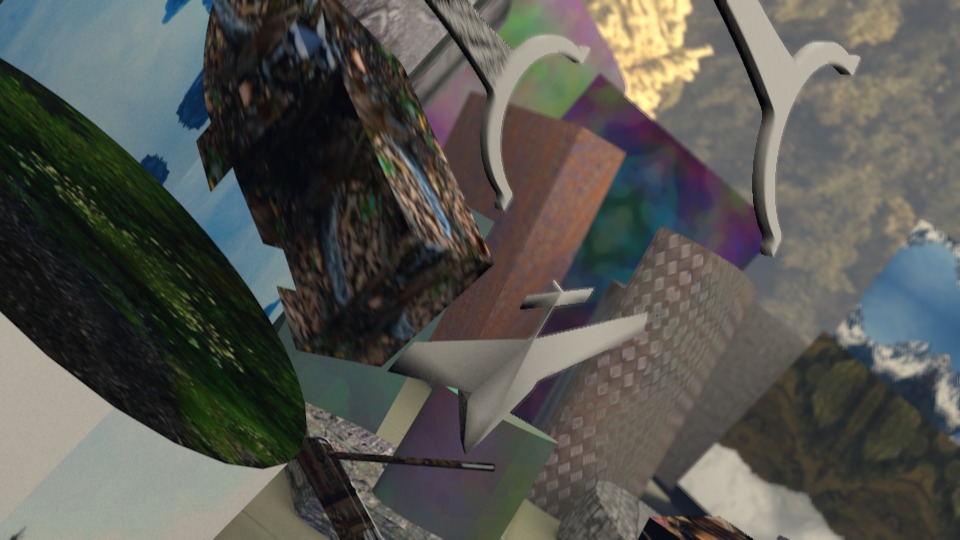} } &
      \multicolumn{2}{c}{ \includegraphics[width=0.49\linewidth]{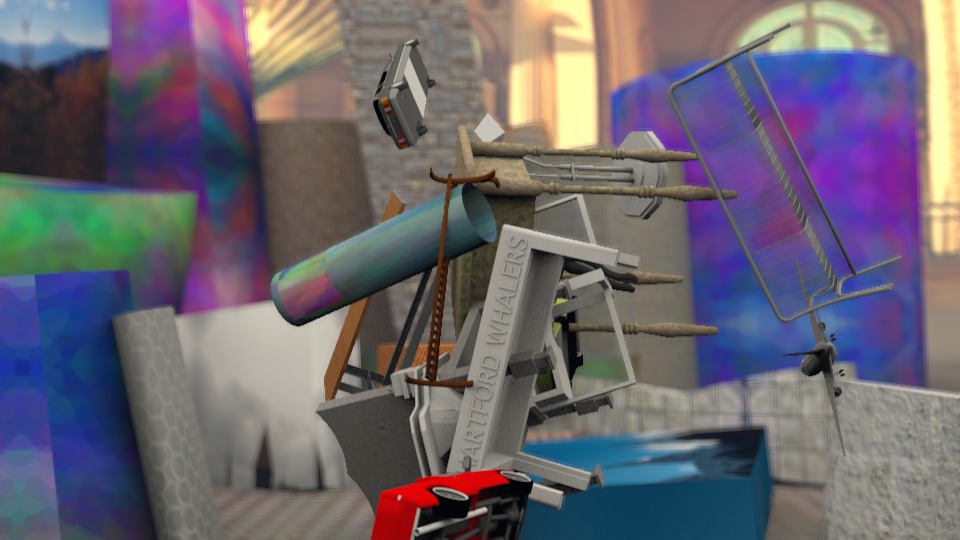} } \\
      
      \includegraphics[width=0.24\linewidth]{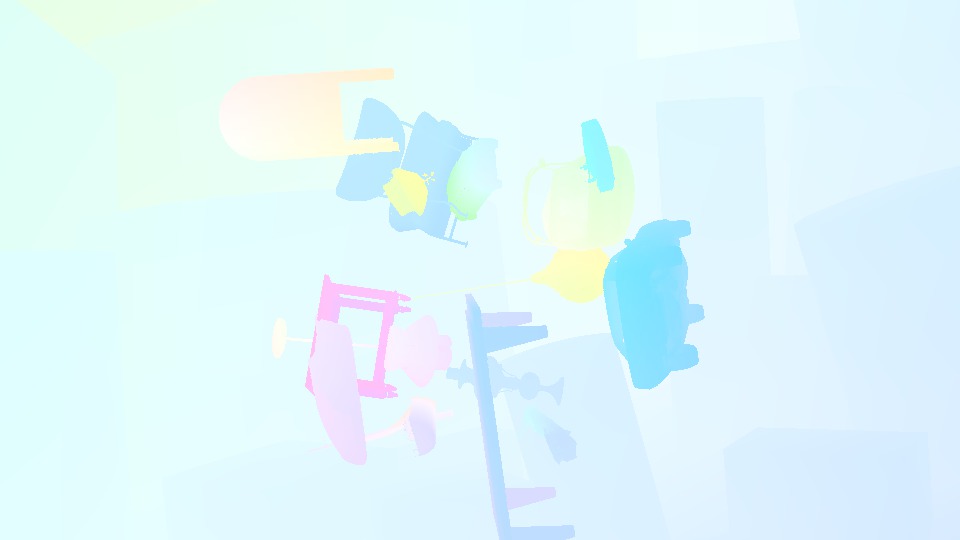} &
      \includegraphics[width=0.24\linewidth]{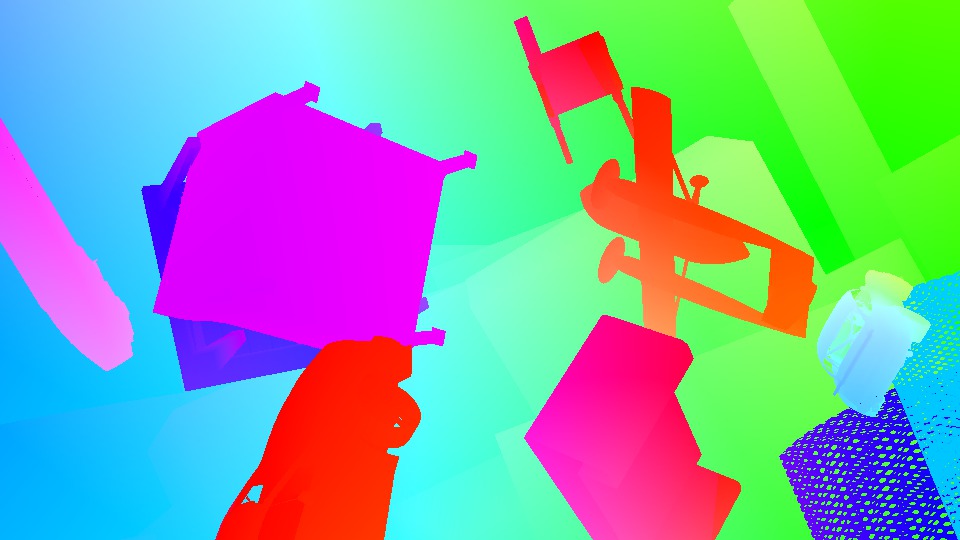} &
      \includegraphics[width=0.24\linewidth]{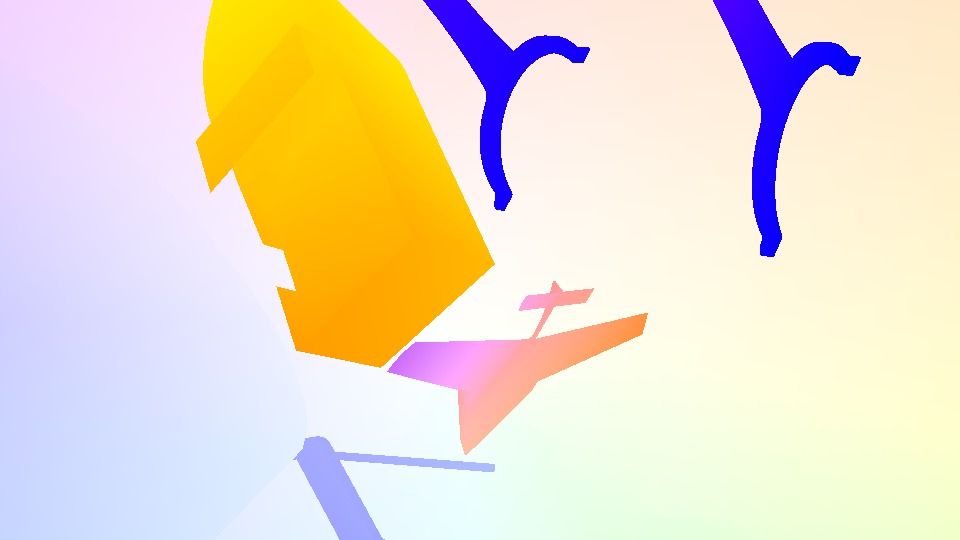} &
      \includegraphics[width=0.24\linewidth]{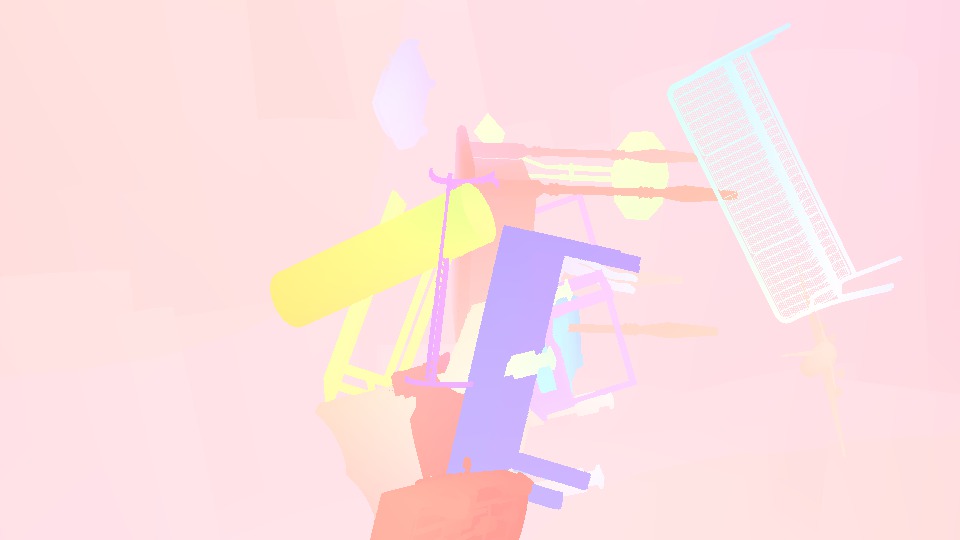} \\
      
      \includegraphics[width=0.24\linewidth]{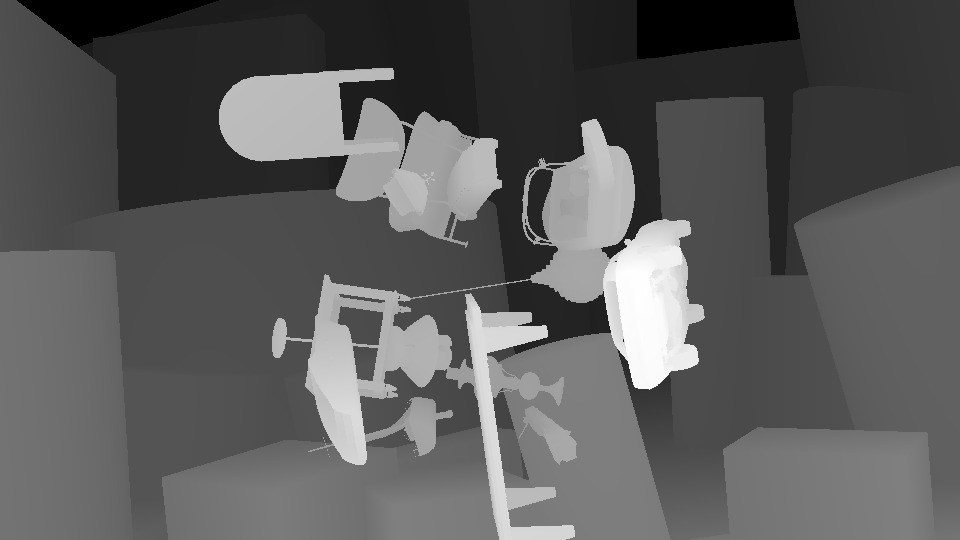} &
      \includegraphics[width=0.24\linewidth]{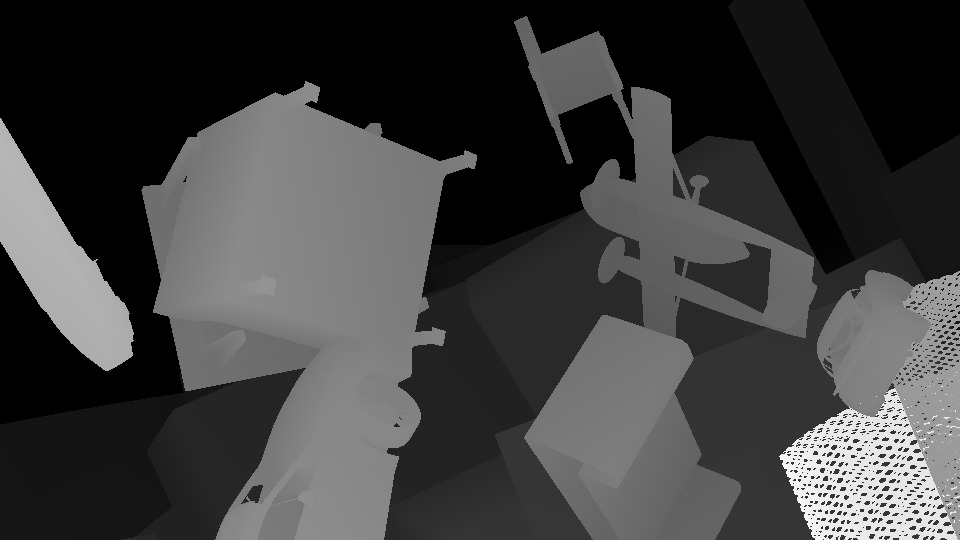} &
      \includegraphics[width=0.24\linewidth]{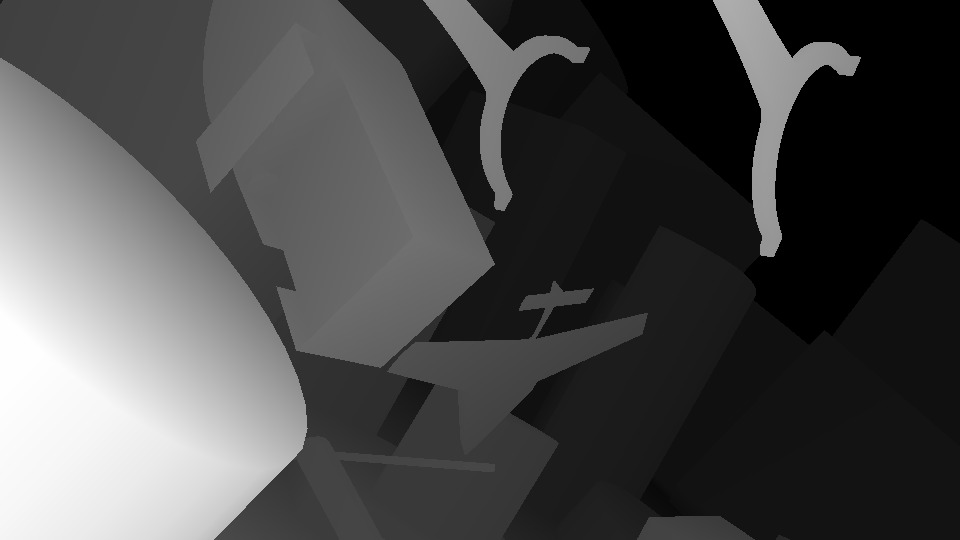} &
      \includegraphics[width=0.24\linewidth]{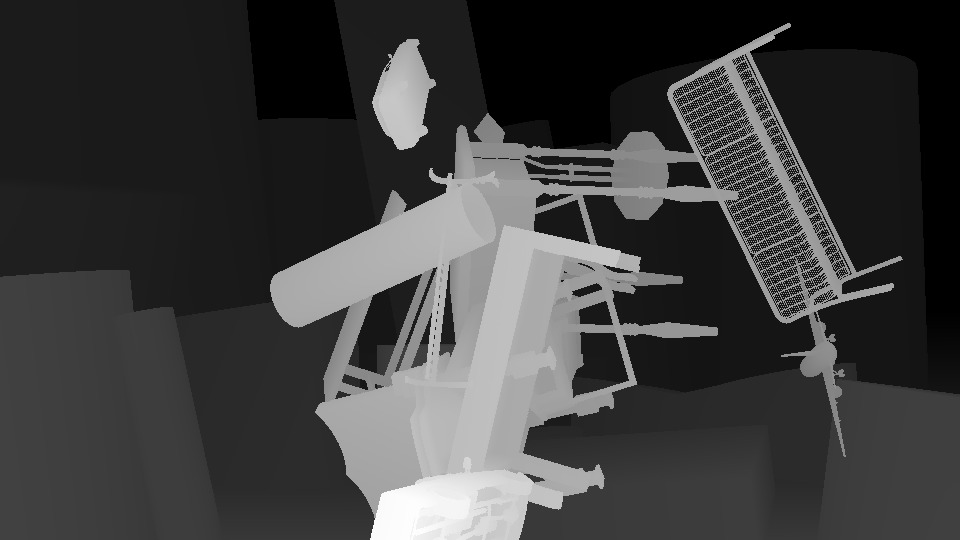} \\
      
      \includegraphics[width=0.24\linewidth]{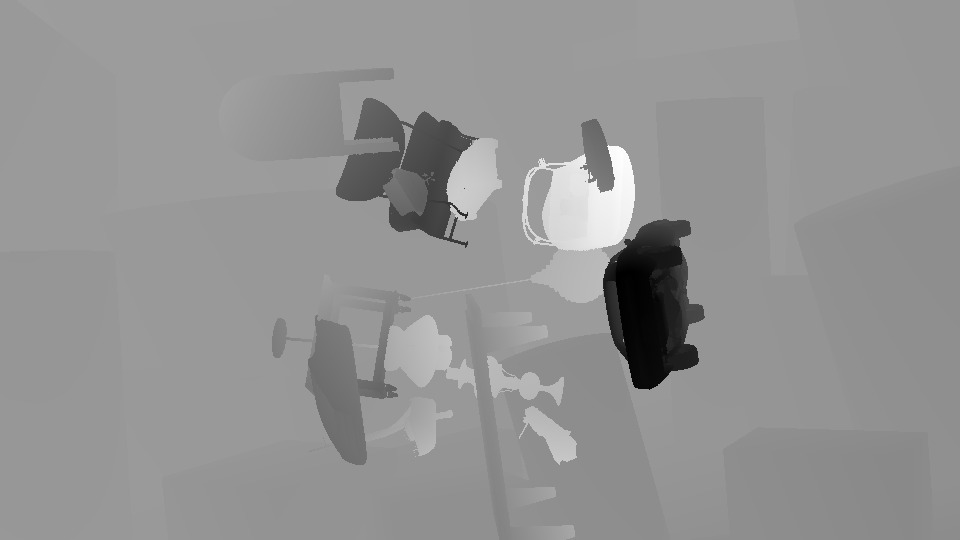} &
      \includegraphics[width=0.24\linewidth]{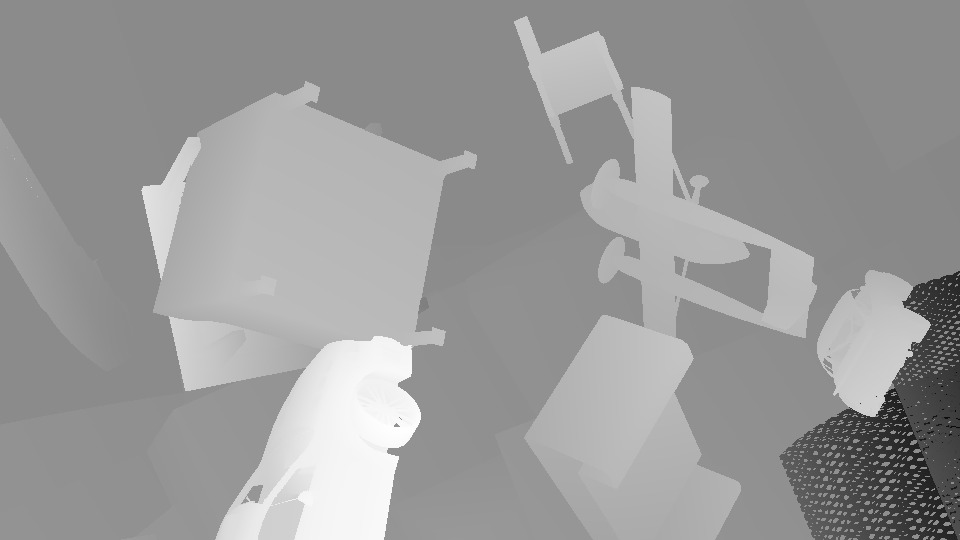} &
      \includegraphics[width=0.24\linewidth]{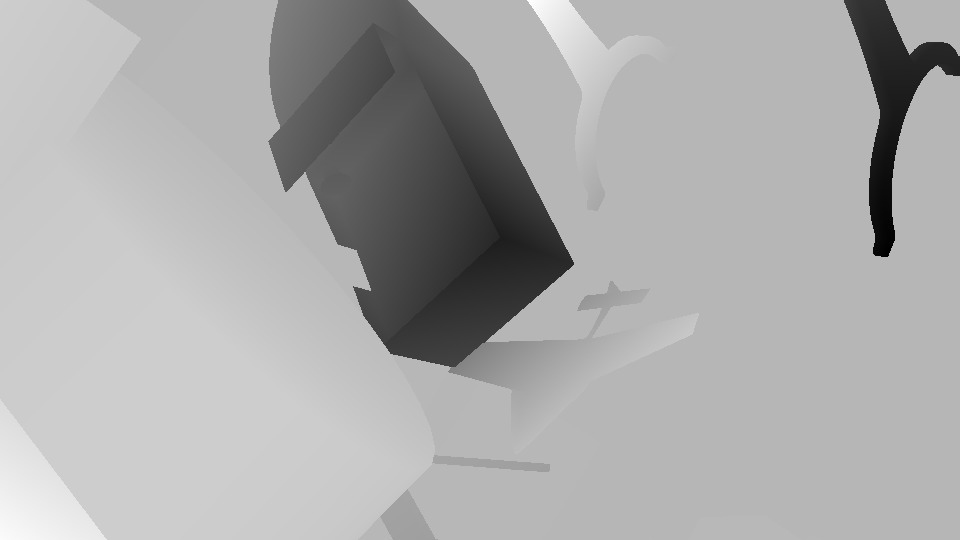} &
      \includegraphics[width=0.24\linewidth]{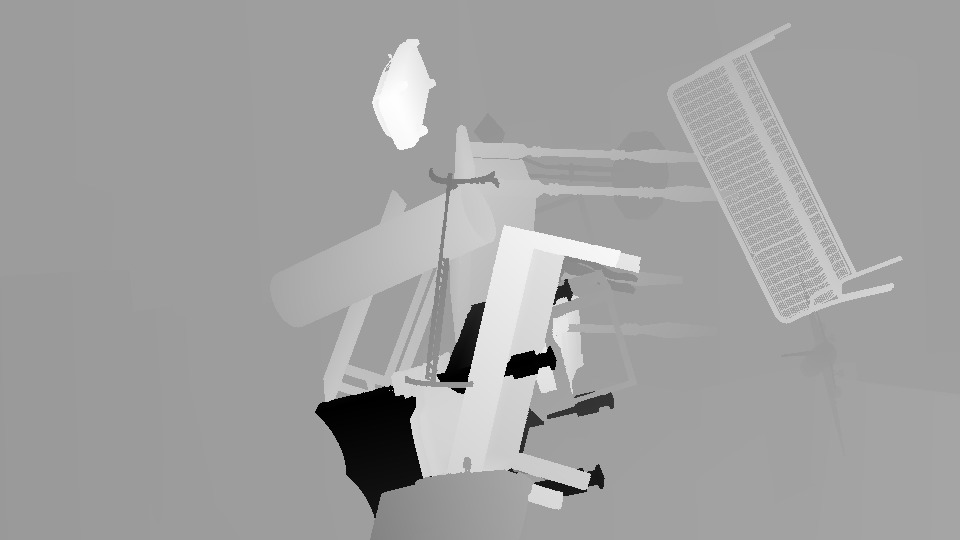} \\
    \end{tabular}
  }
  \end{center}
  \caption{Example scenes from our \emph{FlyingThings3D} dataset.
           {\bf 3rd~row:} Optical flow images,
           {\bf 4th~row:} Disparity images,
           {\bf 5th~row:} Disparity change images.
           Best viewed on a color screen in high resolution (images normalized for display).
          }
  \label{fig:flyingthings}
\end{figure}

The base of each scene is a large textured ground plane. 
We generated $200$ static background objects with shapes that were randomly chosen from cuboids and cylinders.
Each object was randomly scaled, rotated, textured and then placed on the ground plane.

To populate the scene, we downloaded $35\,927$ detailed 3D models from Stanford's ShapeNet~\cite{Savva15}\footnote{\url{http://shapenet.cs.stanford.edu/}} database.
From these we assembled a training set of $32\,872$ models and a testing set of size $3\,055$.
Also the model \emph{categories} were split disjointly.

We sampled between $5$ and $20$ random objects from this object collection and randomly textured every material of every object.
Each ShapeNet object was translated and rotated along a smooth 3D trajectory modeled such that the camera can see the object, but with randomized displacements.
The camera was animated, too.

The texture collection was a combination of procedural images created using ImageMagick\footnote{\url{http://www.imagemagick.org/script/index.php}}, landscape and cityscape photographs from Flickr\footnote{\url{https://www.flickr.com/} Non-commercial public license. We used the code framework by Hays and Efros \cite{Hays-Efros-08}}, and texture-style photographs from Image*After\footnote{\url{http://www.imageafter.com/textures.php}}.
Like the 3D models, also the textures were split into disjoint training and testing parts.

For the final pass images, the scenes vary in presence and intensity of motion blur and defocus blur.

\subsection{Monkaa}\label{sec:monkaa}

The second part of our dataset is made from the open source Blender assets of the animated short film Monkaa\footnote{\url{https://cloud.blender.org/bi/monkaa/}}.
In this regard, it resembles the MPI Sintel dataset.
Monkaa contains nonrigid and softly articulated motion as well as visually challenging fur.
Beyond that, there are few visual similarities to Sintel; the Monkaa movie does not strive for the same amount of naturalism.

We selected a number of suitable movie scenes and additionally created entirely new scenes using parts and pieces from Monkaa.
To increase the amount of data, we rendered our selfmade scenes in multiple versions, each with random incremental changes to the camera's translation and rotation keyframes.


\subsection{Driving}\label{sec:driving}\label{sec:fakekitti}

\begin{figure}[t]
  \begin{center}
  {
    \setlength{\tabcolsep}{1pt}%
    \begin{tabular}{cc}
       \raisebox{8pt}{\includegraphics[width=0.55\linewidth]{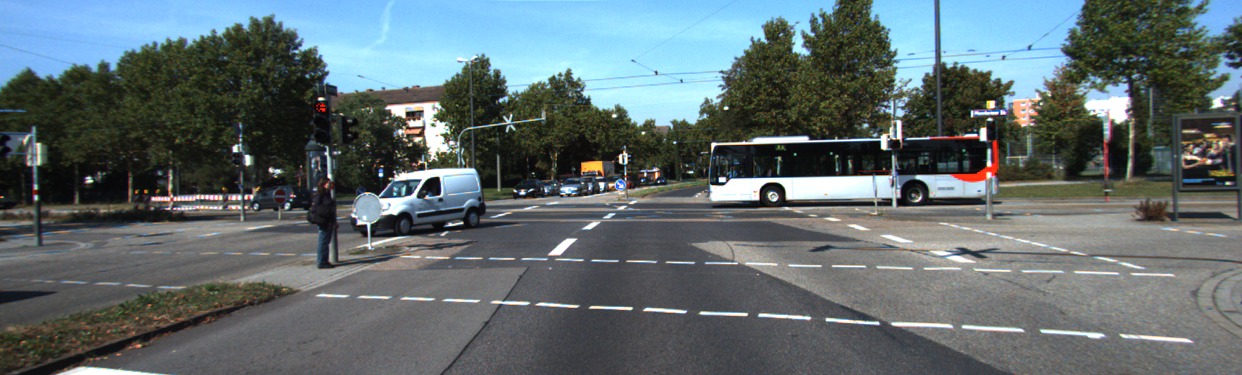}} &
                      \includegraphics[width=0.43\linewidth]{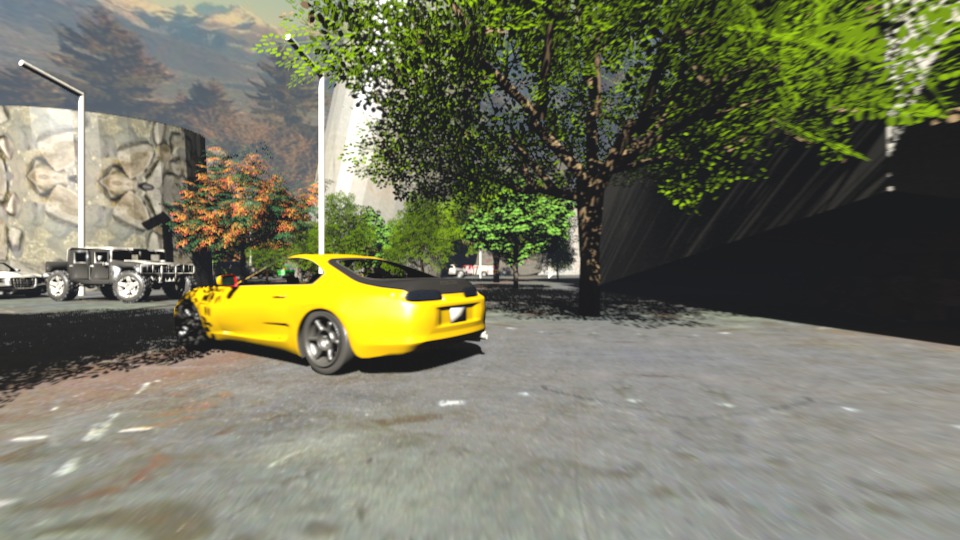} \\
      \raisebox{8pt}{\includegraphics[width=0.55\linewidth]{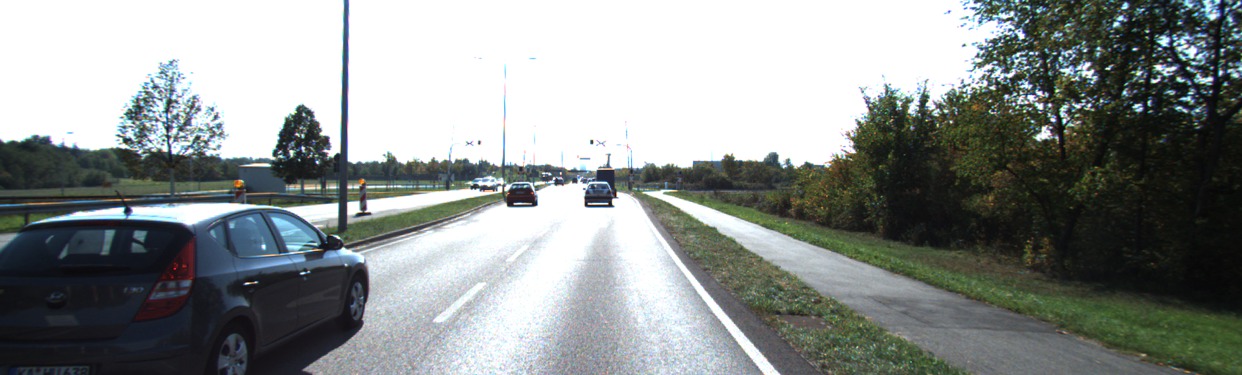}} &
                     \includegraphics[width=0.43\linewidth]{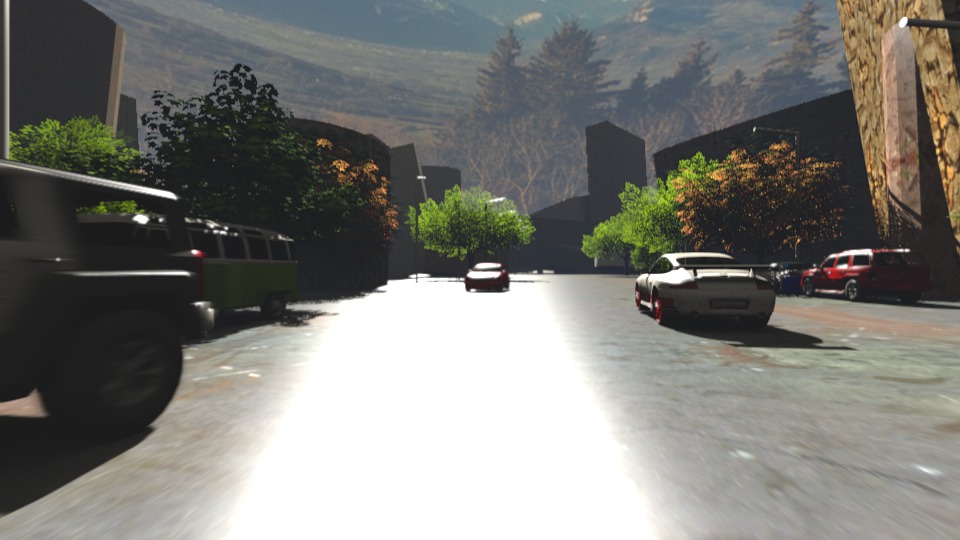} \\
      KITTI 2015 & \emph{Driving} (ours) \\
    \end{tabular}
  }
  \end{center}
  \caption{Example frames from the 2015 version of the KITTI benchmark suite \cite{Menze2015CVPR} and our new \emph{Driving} dataset. Both show many static and moving cars from various realistic viewpoints, thin objects, complex shadows, textured ground, and challenging specular reflections.}
  \label{fig:KITTI_vs_FakeKITTI}
\end{figure}

The \emph{Driving} scene is a mostly naturalistic, dynamic street scene from the viewpoint of a driving car, made to resemble the KITTI datasets.
It uses car models from the same pool as the \emph{FlyingThings3D} dataset and additionally employs highly detailed tree models from 3D Warehouse\footnote{\url{https://3dwarehouse.sketchup.com/}} and simple street lights.
In Fig.~\ref{fig:KITTI_vs_FakeKITTI} we show selected frames from \emph{Driving} and lookalike frames from KITTI 2015.

Our stereo baseline is set to $1$ Blender unit, which together with a typical car model width of roughly $2$ units is comparable to KITTI's setting ($54$cm baseline, $186$cm car width~\cite{Geiger-et-al-13}). 

\section{Networks}\label{sec:nets}
To prove the applicability of our new synthetic datasets to scene flow estimation, we use it to train convolutional networks. In general, we follow the architecture of FlowNet~\cite{FlowNet}.
That is, each network consists of a contractive part and an expanding part with long-range links between them.
The contracting part contains convolutional layers with occasional strides of $2$, resulting in a total downsampling factor of $64$.
This allows the network to estimate large displacements.
The expanding part of the network then gradually and nonlinearly upsamples the feature maps, taking into account also the features from the contractive part.
This is done by a series of up-convolutional and convolutional layers. Note that there is no data bottleneck in the network, as information can also pass through the long-range connections between contracting and expanding layers.
For an illustration of the overall architecture we refer to the figures in Dosovitskiy et al.~\cite{FlowNet}.

For disparity estimation we propose the basic architecture \emph{DispNet} described in Table~\ref{table:archs}. We found that additional convolutions in the expanding part yield smoother disparity maps than the FlowNet architecture; see Figure~\ref{fig:interconv}.

We also tested an architecture that makes use of an explicit correlation layer~\cite{FlowNet}, which we call \emph{DispNetCorr}.
In this network, the two images are processed separately up to layer conv2 and the resulting features are then correlated horizontally. We consider a maximum displacement of $40$ pixels, which corresponds to $160$ pixels in the input image. 
Compared to the 2D correlation in Dosovitskiy et al.~\cite{FlowNet}, 1D correlation is computationally much cheaper and allows us to cover larger displacements with finer sampling than in the FlowNet, which used a stride of $2$ for the correlation. 

\begin{figure}[t]
\begin{center}
  \includegraphics[width=0.9\linewidth]{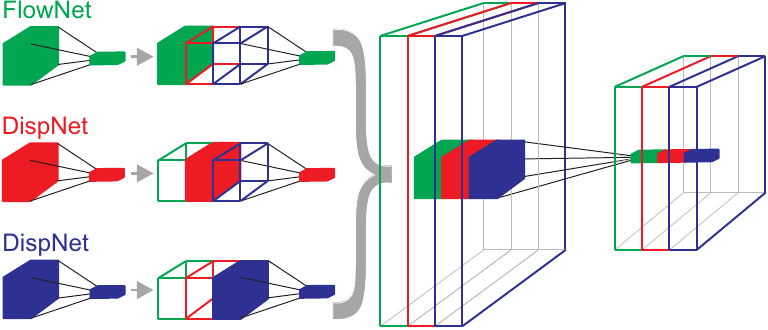} \\
  \end{center}
  \caption{Interleaving the weights of a FlowNet (green) and two DispNets (red and blue) to a SceneFlowNet. For every layer, the filter masks are created by taking the weights of one network (left) and setting the weights of the other networks to zero, respectively (middle). The outputs from each network are then concatenated to yield one big network with three times the number of inputs and outputs (right).}
  \label{fig:interleaving}
\end{figure}

We train a joint network for scene flow estimation by combining and fine-tuning pretrained networks for disparity and flow.
This is illustrated in Figure~\ref{fig:interleaving}. We use our implementation of FlowNet to predict flow between the left and right image and two DispNets to predict the disparities at $t$ and $t\!+\!1$. We then fine-tune the large combined network to estimate flow, disparity, and additionally disparity change. 

\begin{table}
   \begin{center}
  \resizebox{0.98\linewidth}{!}{%
  \begin{tabular}{|l|ccc|cc|c|}
      \hline
      Name & Kernel & Str. & Ch I/O & InpRes & OutRes & Input\\
      \hline 
      conv1 & $7\!\times\!7$ & 2 & $6/64$ & $768\!\times\!384$ & $384\!\times\!192$ & Images \\
      conv2 & $5\!\times\!5$ & 2 & $64/128$ & $384\!\times\!192$ & $192\!\times\!96$ & conv1 \\
      conv3a & $5\!\times\!5$ & 2 & $128/256$ & $192\!\times\!96$ & $96\!\times\!48$ & conv2 \\
      conv3b & $3\!\times\!3$ & 1 & $256/256$ & $96\!\times\!48$ & $96\!\times\!48$ & conv3a \\
      conv4a & $3\!\times\!3$ & 2 & $256/512$ & $96\!\times\!48$ & $48\!\times\!24$ & conv3b \\
      conv4b & $3\!\times\!3$ & 1 & $512/512$ & $48\!\times\!24$ & $48\!\times\!24$ & conv4a \\
      conv5a & $3\!\times\!3$ & 2 & $512/512$ & $48\!\times\!24$ & $24\!\times\!12$ & conv4b \\
      conv5b & $3\!\times\!3$ & 1 & $512/512$ & $24\!\times\!12$ & $24\!\times\!12$ & conv5a \\
      conv6a & $3\!\times\!3$ & 2 & $512/1024$ & $24\!\times\!12$ & $12\!\times\!6$ & conv5b \\
      conv6b & $3\!\times\!3$ & 1 & $1024/1024$ & $12\!\times\!6$ & $12\!\times\!6$ & conv6a \\
      \hline
      pr6+loss6 & $3\!\times\!3$ & 1 & $1024/1$ & $12\!\times\!6$ & $12\!\times\!6$ & conv6b \\
      \hline
      upconv5 & $4\!\times\!4$ & 2 & $1024/512$ & $12\!\times\!6$ & $24\!\times\!12$ & conv6b \\
      iconv5 & $3\!\times\!3$ & 1 & $1025/512$ & $24\!\times\!12$ & $24\!\times\!12$ & upconv5+pr6+conv5b \\
      pr5+loss5 & $3\!\times\!3$ & 1 & $512/1$ & $24\!\times\!12$ & $24\!\times\!12$ & iconv5 \\
      upconv4 & $4\!\times\!4$ & 2 & $512/256$ & $24\!\times\!12$ & $48\!\times\!24$ & iconv5 \\
      iconv4 & $3\!\times\!3$ & 1 & $769/256$ & $48\!\times\!24$ & $48\!\times\!24$ & upconv4+pr5+conv4b \\
      pr4+loss4 & $3\!\times\!3$ & 1 & $256/1$ & $48\!\times\!24$ & $48\!\times\!24$ & iconv4 \\
      upconv3 & $4\!\times\!4$ & 2 & $256/128$ & $48\!\times\!24$ & $96\!\times\!48$ & iconv4 \\
      iconv3 & $3\!\times\!3$ & 1 & $385/128$ & $96\!\times\!48$ & $96\!\times\!48$ & upconv3+pr4+conv3b \\
      pr3+loss3 & $3\!\times\!3$ & 1 & $128/1$ & $96\!\times\!48$ & $96\!\times\!48$ & iconv3 \\
      upconv2 & $4\!\times\!4$ & 2 & $128/64$ & $96\!\times\!48$ & $192\!\times\!96$ & iconv3 \\
      iconv2 & $3\!\times\!3$ & 1 & $193/64$ & $192\!\times\!96$ & $192\!\times\!96$ & upconv2+pr3+conv2 \\
      pr2+loss2 & $3\!\times\!3$ & 1 & $64/1$ & $192\!\times\!96$ & $192\!\times\!96$ & iconv2 \\
      upconv1 & $4\!\times\!4$ & 2 & $64/32$ & $192\!\times\!96$ & $384\!\times\!192$ & iconv2 \\
      iconv1 & $3\!\times\!3$ & 1 & $97/32$ & $384\!\times\!192$ & $384\!\times\!192$ & upconv1+pr2+conv1 \\
      pr1+loss1 & $3\!\times\!3$ & 1 & $32/1$ & $384\!\times\!192$ & $384\!\times\!192$ & iconv1 \\
      \hline
    \end{tabular}}
  \end{center}
  \caption{Specification of DispNet architecture. The contracting part consists of convolutions \emph{conv1} to \emph{conv6b}. In the expanding part, upconvolutions (\emph{upconvN}), convolutions (\emph{iconvN}, \emph{prN}) and loss layers are alternating. Features from earlier layers are concatenated with higher layer features. The predicted disparity image is output by \emph{pr1}.}
  \label{table:archs}
\end{table}

\begin{table*}
  \begin{center}
    \resizebox{0.98\textwidth}{!}{%
    \begin{tabular}{|l||cc|cc|c|c|c|c||l|}
      \hline
      Method & \multicolumn{2}{|c|}{KITTI 2012} & \multicolumn{2}{|c|}{KITTI 2015} & Driving & FlyingThings3D & Monkaa & Sintel Clean & Time\\
             & train & test & train & test (D1) &  & test & & train & \\
      \hline
      \hline
      DispNet         & $2.38$  & ---   & $2.19$   & ---       & $15.62$  & $2.02$ & $5.99$ & $5.38$ & $0.06$s\\ 
      DispNetCorr1D   & $1.75$  & ---   & $1.59$   & ---       & $16.12$ & $1.68$ & $5.78$ & $5.66$ & $0.06$s \\ 
      DispNet-K       & $1.77$  & ---   & $(0.77)$ & ---       & $19.67$  & $7.14$ & $14.09$ & $21.29$ & $0.06$s \\ 
      DispNetCorr1D-K & $1.48$  & $1.0^\dagger$ & $(0.68)$ & $4.34\%$  & $20.40$  & $7.46$ & $14.93$ & $21.88$ & $0.06$s \\ 
      SGM             & $10.06$ & ---   & $7.21$   & $10.86\%$ & $40.19$  & $8.70$ & $20.16$ & $19.62$ & 1.1s \\
      MC-CNN-fst      & ---     & ---   & ---      & $4.62\%$  & $19.58$ & $4.09$ & $6.71$ & $11.94$ & $0.8$s \\
      MC-CNN-acrt     & ---     & $0.9$ & ---      & $3.89\%$  & ---      & ---    & --- & --- & $67$s \\
      \hline
      \end{tabular}}
  \end{center}
  \caption{Disparity errors. All measures are endpoint errors, except for the \emph{D1-all} measure (see the text for explanation) for KITTI-2015 test. ${}^\dagger$ This result is from a network fine-tuned on KITTI 2012 train.}
  \label{table:evaluation}
\end{table*}

\textbf{Training.}
All networks are trained end-to-end, given the images as input and the ground truth (optical flow, disparity, or scene flow) as output.
We employ a custom version of Caffe~\cite{caffe} and make use of the Adam optimizer~\cite{Kingma-14}.
We set $\beta_1=0.9$ and $\beta_2=0.999$ as in Kingma et al.~\cite{Kingma-14}. As learning rate we used $\lambda=1e-4$ and divided it by $2$ every $200\,000$ iterations starting from iteration $400\,000$.

Due to the depth of the networks and the direct connections between contracting and expanding layers (see Table~\ref{table:archs}), lower layers get mixed gradients if all six losses are active. We found that using a loss weight schedule can be beneficial: we start training with a loss weight of 1 assigned to the lowest resolution loss \emph{loss6} and a weight of 0 for all other losses (that is, all other losses are switched off). During training, we progressively increase the weights of losses with higher resolution and deactivate the low resolution losses. This enables the network to first learn a coarse representation and then proceed with finer resolutions without losses constraining intermediate features.

\textbf{Data Augmentation.}
Despite the large training set, we chose to perform data augmentation to introduce more diversity into the training data at almost no extra cost\footnote{The computational bottleneck is in reading the training samples from disk, whereas data augmentation is performed on the fly.}.
We perform spatial transformations (rotation, translation, cropping, scaling) and chromatic transformations (color, contrast, brightness), and we use the same transformation for all 2 or 4 input images.

For disparity, introducing any rotation or vertical shift would break the epipolar constraint. Horizontal shifts would lead to negative disparities or shifting infinity towards the camera.

\section{Experiments}\label{sec:experiments}
\paragraph{Evaluation of existing methods.}
We evaluated several existing disparity and optical flow estimation methods on our new dataset.
Namely, for disparity we evaluate the state-of-the-art method of {\v{Z}}bontar and 
LeCun~\cite{zbontar2015stereo} and the popular Semi-Global Matching~\cite{Hirschmueller2008PAMI} 
approach with a block matching implementation from OpenCV\footnote{\url{http://docs.opencv.org/2.4/modules/calib3d/doc/camera_calibration_and_3d_reconstruction.html\#stereosgbm}}.
Results are shown together with those of our DispNets in Table~\ref{table:evaluation}.
We use the endpoint error (EPE) as error measure in most cases, with the only exception of KITTI 2015 test set where only the \emph{D1-all} error measure is reported by the KITTI evaluation server.
It is the percentage of pixels for which the estimation error is larger than $3$px and larger than $5 \%$ of the ground truth disparity at this pixel.

\paragraph{DispNet.}
We train DispNets on the FlyingThings3D dataset and then optionally fine-tune it on KITTI.
The fine-tuned networks are denoted by '-K' suffix in the table.
DispNetCorr fine-tuned on KITTI 2015 is currently second best in the KITTI 2015 top results table, slightly behind MC-CNN-acrt~\cite{zbontar2015stereo}, while being roughly $1000$ times faster. On KITTI resolution it runs at 15 frames per second on an Nvidia GTX TitanX GPU. 
For foreground pixels (belonging to car models) it reaches an error that is roughly half the error of~\cite{zbontar2015stereo}.
The network achieves about $30 \%$ lower error than the best real-time method reported in the table, Multi-Block-Matching~\cite{EineckeE15}.
Also on the other datasets DispNet performs well and outperforms both SGM and MC-CNN. 

While fine-tuning on KITTI improves the results on this dataset, it increases errors on other datasets.
We explain this significant performance drop by the fact that KITTI 2015 only contains relatively small disparities, up to roughly $150$ pixels, while the other datasets contain some disparities of $500$ pixels and more. 
When fine-tuned on KITTI, the network seems to lose its ability to predict large displacements, hence making huge errors on these.

\begin{figure}
\includegraphics[width=\linewidth]{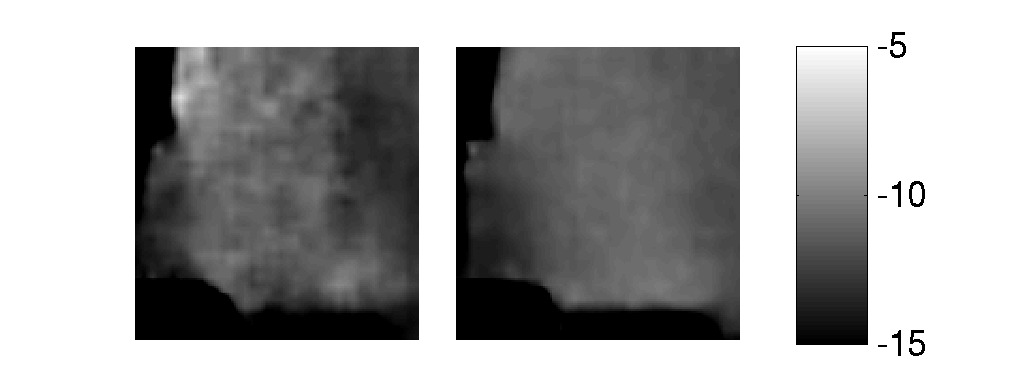}
\caption{Close-up of a predicted disparity map without (\textbf{left}) and with (\textbf{right}) convolutions between up-convolutions. Note how the prediction on the right is much smoother.}
\label{fig:interconv}
\end{figure}

We introduced several modifications to the network architecture compared to the FlowNet~\cite{FlowNet}.
First, we added convolutional layers between up-convolutional layers in the expanding part of the network. 
As expected, this allows the network to better regularize the disparity map and predict smoother results, as illustrated in Figure~\ref{fig:interconv}. 
Quantitatively, this results in roughly $15 \%$ relative EPE decrease on KITTI-2015.

Second, we trained a version of our network with a 1D correlation layer. 
In contrast to Dosovitskiy et al.~\cite{FlowNet}, we find that networks with correlation are systematically better (see Table~\ref{table:evaluation}).
A likely plausible explanation is that the 1D nature of the disparity estimation problem allows us to compute correlations at a finer grid than the FlowNet. 

\begin{figure*}[t]
  \begin{center}
  {
    \setlength{\tabcolsep}{1pt}%
    \begin{tabular}{ccccc}
      RGB image 0/1 (L) & 
      RGB image 0/1 (R) & 
      flow GT / prediction & 
      disp GT / prediction & 
      $\Delta$disp GT / prediction \\
      
      \includegraphics[width=0.195\linewidth]{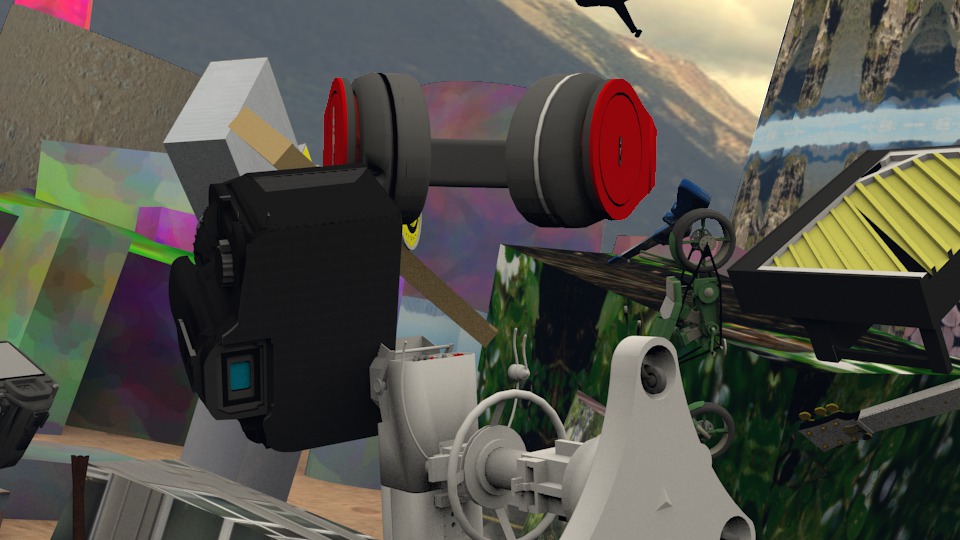} &
      \includegraphics[width=0.195\linewidth]{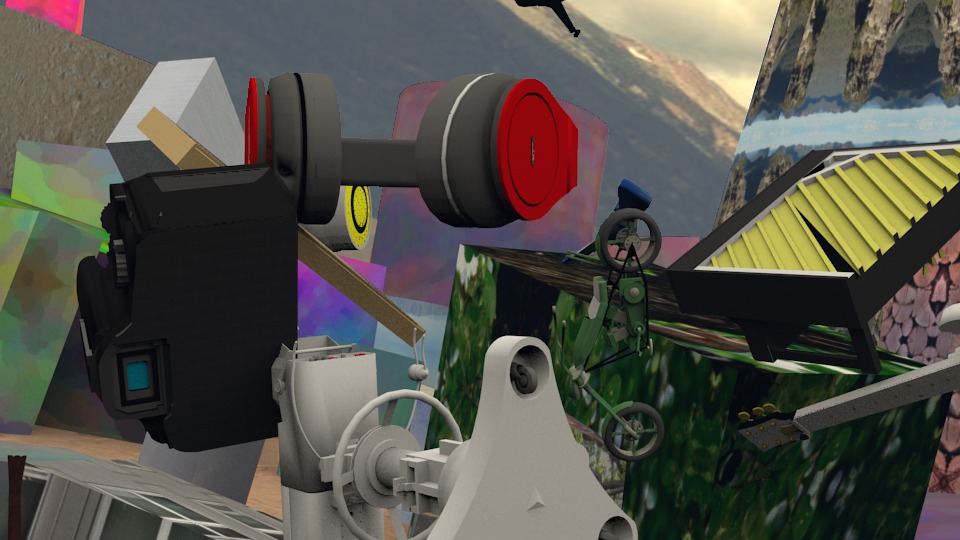} &
      \includegraphics[width=0.195\linewidth]{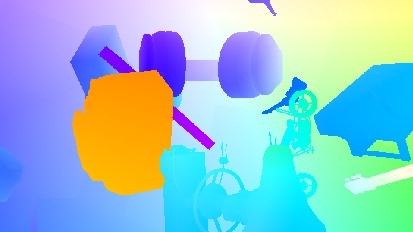} &
      \includegraphics[width=0.195\linewidth]{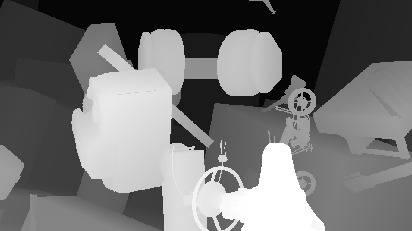} &
      \includegraphics[width=0.195\linewidth]{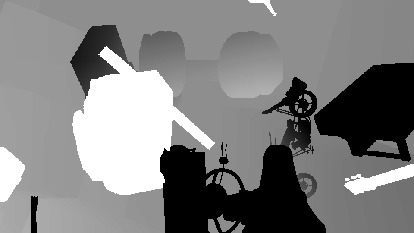} \\
      
      \includegraphics[width=0.195\linewidth]{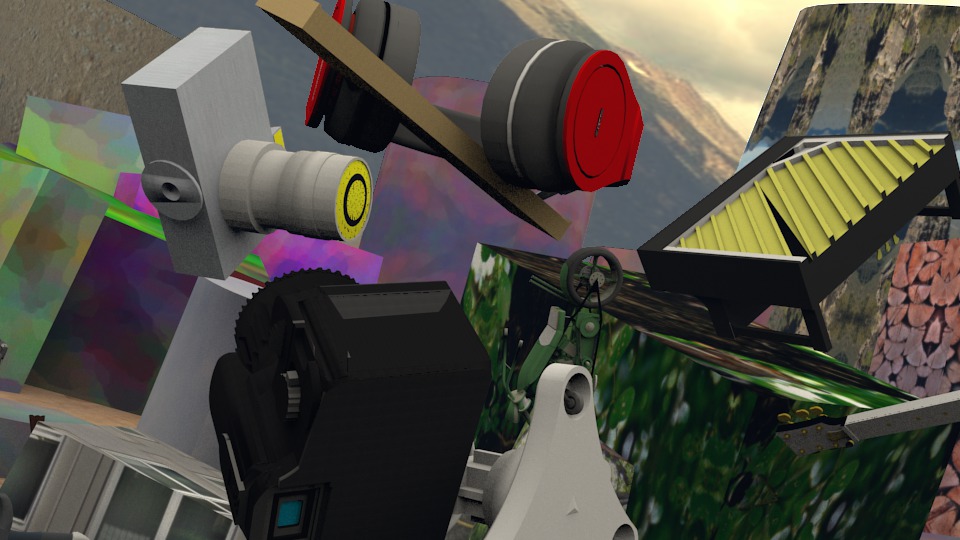} &
      \includegraphics[width=0.195\linewidth]{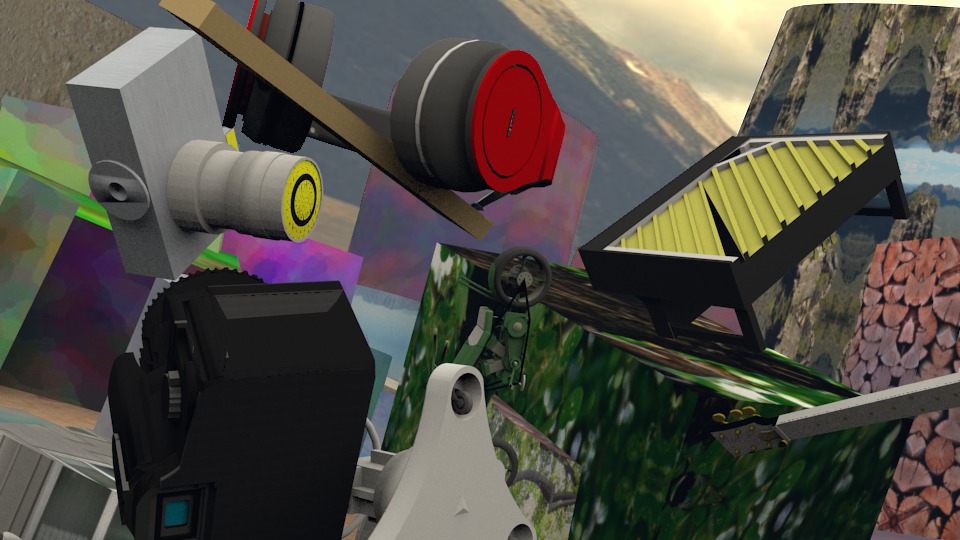} &
      \includegraphics[width=0.195\linewidth]{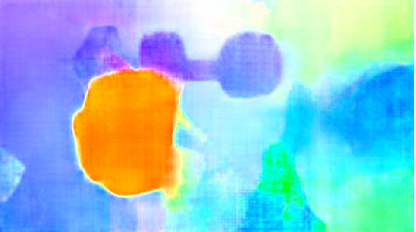} &
      \includegraphics[width=0.195\linewidth]{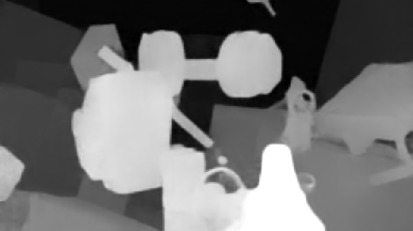} &
      \includegraphics[width=0.195\linewidth]{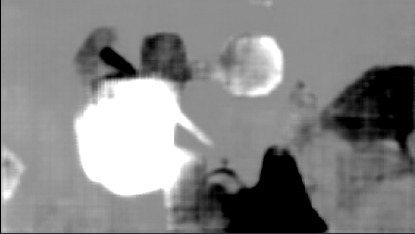} \\
    \end{tabular}
  }
  \end{center}
  \caption{Results of our SceneFlowNet created from pretrained FlowNet and DispNets. The disparity change was added and the network was fine-tuned on FlyingThings3D for $80\,000$ iterations. The disparity change predictions are already quite good after these few training iterations.}
  \label{fig:gallery_sceneflow}
\end{figure*}

\paragraph{SceneFlowNet.}
We present early results on full scene flow estimation with a convolutional network.
Figure~\ref{fig:gallery_sceneflow} shows the results of the network for one FlyingThings scene. 
The network is able to predict disparity change  well, even in the regions which get occluded.
Due to the large amount of data that has to be processed when training for scene flow, the network training is relatively slow (a forward pass of the network takes 0.28s, $5$ times longer than on a DispNet) and has not converged yet. 
We expect the results to further improve as we allow the network to train longer.  
A quantitative evaluation on our datasets is shown in Table~\ref{table:sceneflownet_evaluation}. 

\begin{table}
  \begin{center}
    \begin{tabular}{|l||c|c|c|}
      \hline
      SceneFlowNet & Driving & FlyingThings3D & Monkaa \\
      \hline
      \hline
      Flow              & $22.01$  & $13.45$ & $7.68$ \\
      Disparity         & $17.56$  & $2.37$ & $6.16$ \\
      Disp. change  & $16.89$  & $0.91$ & $0.81$ \\
      \hline
      \end{tabular}
  \end{center}
  \caption{Endpoint errors for the evaluation of our SceneFlowNet on the presented datasets. 
  The Driving dataset contains the largest disparities, flows and disparity changes, resulting in large errors. 
  The FlyingThings3D dataset contains large flows, while Monkaa contains smaller flows and larger disparities.}
  \label{table:sceneflownet_evaluation}
\end{table}


\begin{figure*}[t]
  \begin{center}
  {
    \setlength{\tabcolsep}{1pt}%
    \begin{tabular}{ccccc}
      RGB image (L) & 
      DispNetCorr1D-K &
      MC-CNN prediction &
      SGM prediction \\

      \includegraphics[width=0.245\linewidth]{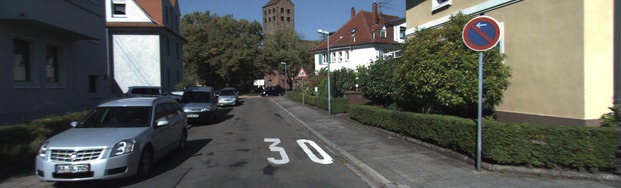} &
      \includegraphics[width=0.245\linewidth]{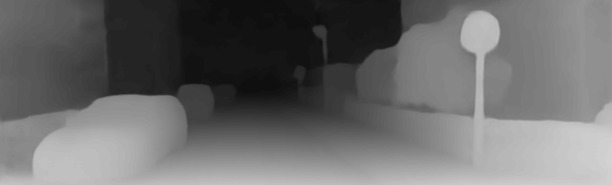} &
      \includegraphics[width=0.245\linewidth]{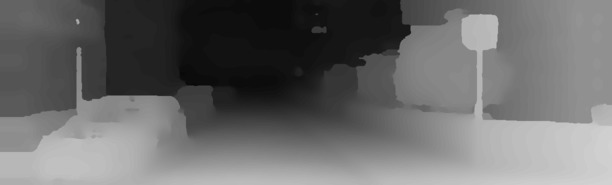} &
      \includegraphics[width=0.245\linewidth]{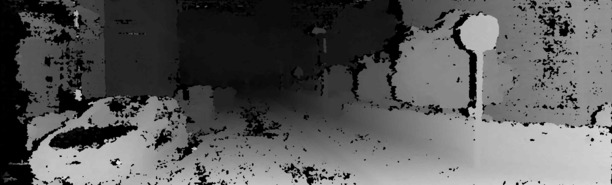} \\
      
      \includegraphics[width=0.245\linewidth]{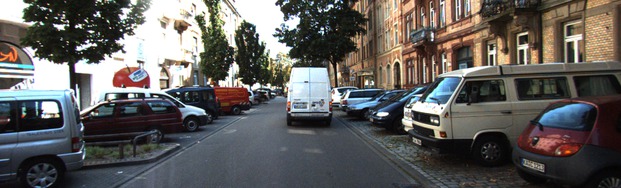} &
      \includegraphics[width=0.245\linewidth]{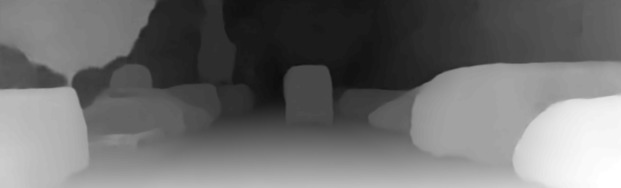} &
      \includegraphics[width=0.245\linewidth]{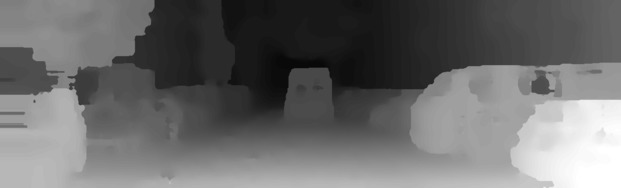} &
      \includegraphics[width=0.245\linewidth]{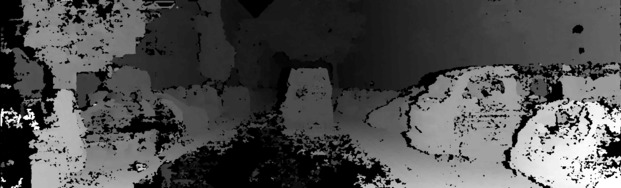} \\

    \end{tabular}
    \begin{tabular}{ccccc}
      RGB image (L) & 
      Disparity GT & 
      DispNetCorr1D &
      MC-CNN prediction &
      SGM prediction \\
      
      \includegraphics[width=0.195\linewidth]{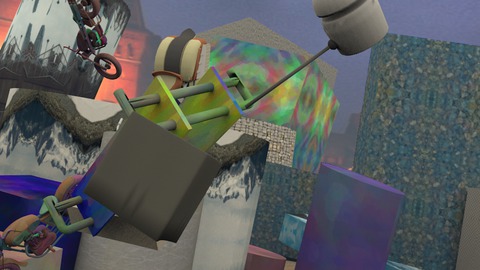} &
      \includegraphics[width=0.195\linewidth]{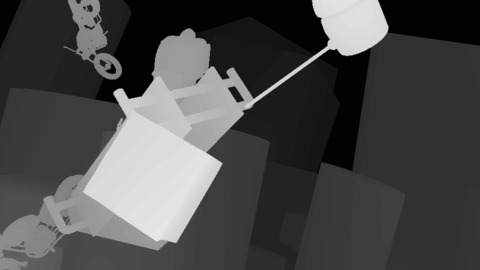} &
      \includegraphics[width=0.195\linewidth]{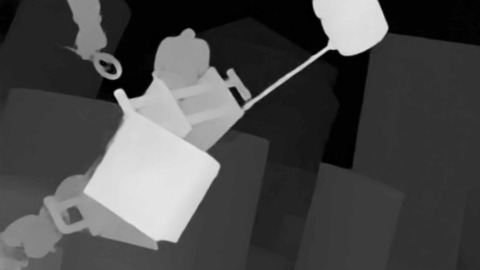} &
      \includegraphics[width=0.195\linewidth]{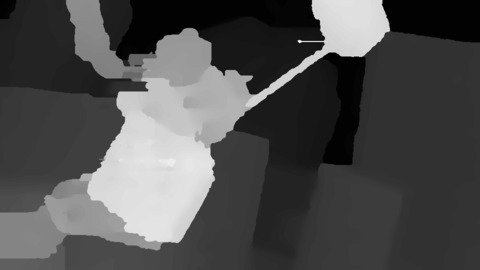} &
      \includegraphics[width=0.195\linewidth]{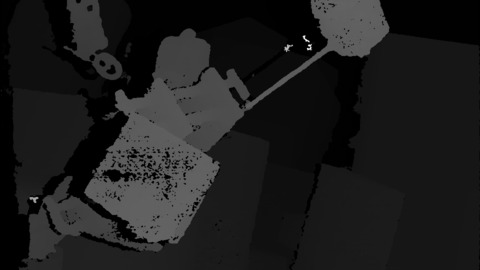} \\
      
      \includegraphics[width=0.195\linewidth]{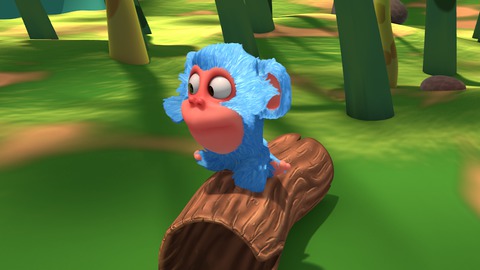} &
      \includegraphics[width=0.195\linewidth]{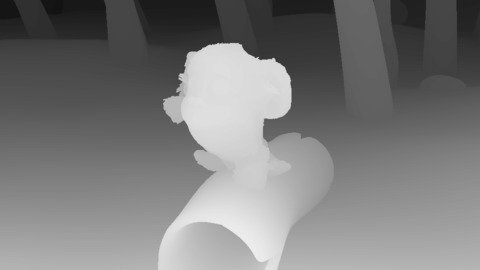} &
      \includegraphics[width=0.195\linewidth]{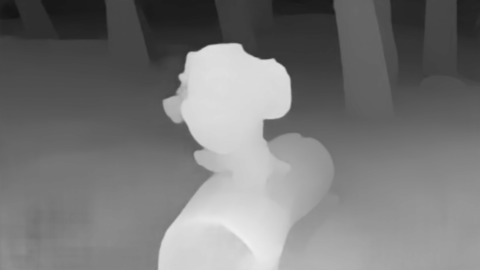} &
      \includegraphics[width=0.195\linewidth]{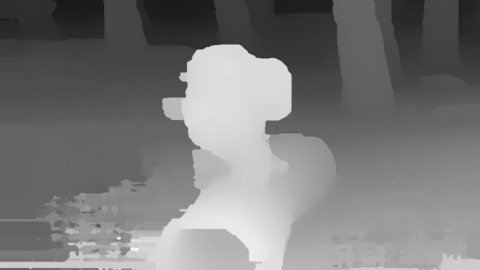} &
      \includegraphics[width=0.195\linewidth]{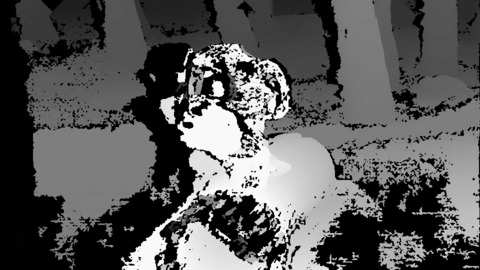} \\
      
      \includegraphics[width=0.195\linewidth]{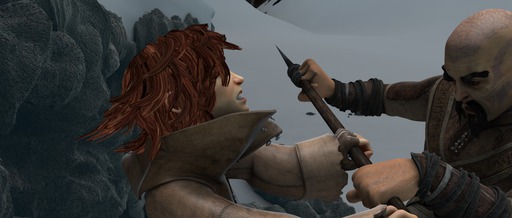} &
      
      \includegraphics[width=0.195\linewidth]{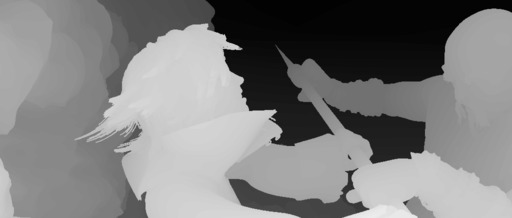} &
      \includegraphics[width=0.195\linewidth]{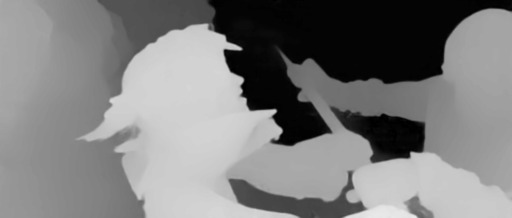} &
      \includegraphics[width=0.195\linewidth]{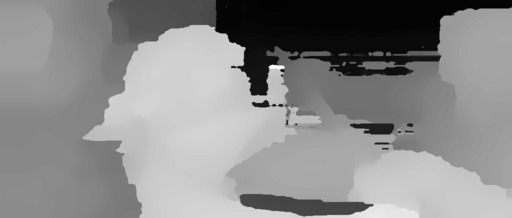} &
      \includegraphics[width=0.195\linewidth]{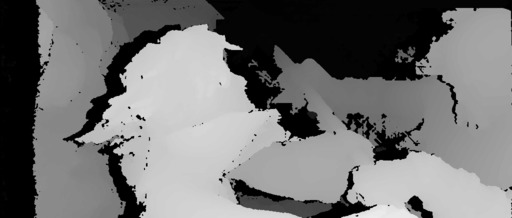} \\
    \end{tabular}
  }
  \end{center}
  \caption{Disparity Results. Rows from top to bottom: KITTI 2012, KITTI 2015, FlyingThings3D, Monkaa, Sintel. Note how the DispNet prediction is basically noise-free.}
  \label{fig:gallery_disp}
\end{figure*}

\section{Conclusion}\label{sec:conclusion}
We have introduced a  synthetic dataset containing over $35\,000$ stereo image pairs with ground truth disparity, optical flow, and scene flow. While our motivation was to create a sufficiently large dataset that is suitable to train convolutional networks to estimate these quantities, the dataset can also serve for evaluation of other methods. This is particularly interesting for scene flow, where there has been a lack of datasets with ground truth.

We have demonstrated that the dataset can indeed be used to successfully train large convolutional networks: the network we trained for disparity estimation is on par with the state of the art and runs 1000 times faster. 
A first approach of training the network for scene flow estimation using a standard network architecture also shows promising results.  We are convinced that our dataset will help to boost deep learning research for such challenging vision tasks as stereo, flow and scene flow estimation.


\section{Acknowledgements}\label{sec:acknowledgements}
The work was partially funded by the ERC Starting Grant VideoLearn, the ERC Consolidator Grant 3D Reloaded, and by the DFG Grants \mbox{BR 3815/7-1} and \mbox{CR 250/13-1}.

{\small
\bibliographystyle{ieee}
\bibliography{bibliography}
}

\begin{figure*}%
  \begin{center}%
    \textbf{\Large A Large Dataset to Train Convolutional Networks for Disparity, Optical Flow,\\\vspace{.25cm} and Scene Flow Estimation: Supplementary Material}
  \end{center}%
\end{figure*}%


\pagebreak

\setcounter{section}{0}
\setcounter{figure}{0}
\setcounter{footnote}{0}

\begin{figure}%
  \begin{center}%
  {%
    \resizebox{0.98\linewidth}{!}{%
      \includegraphics[width=.2\linewidth]{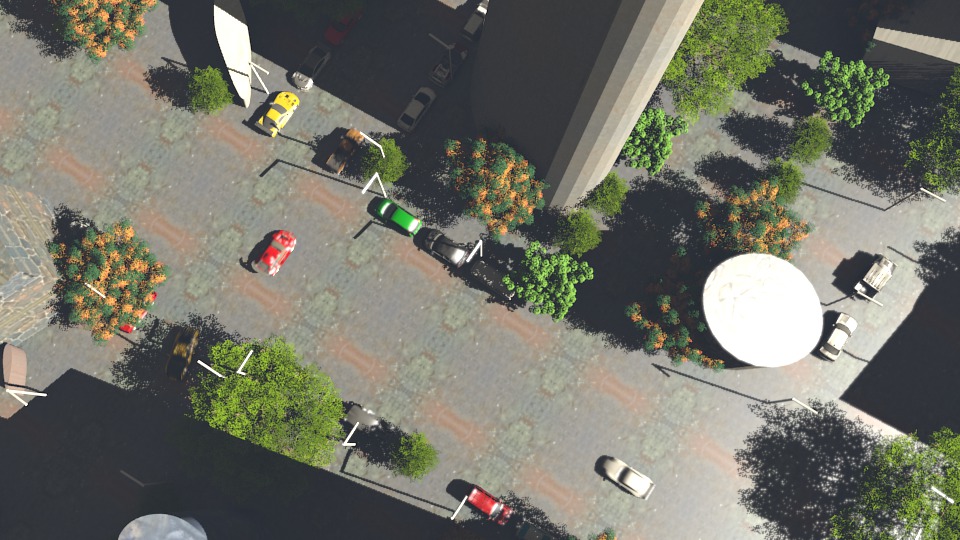}%
    }%
  }%
  \end{center}%
  \caption{Bird's eye view of the \emph{Driving} scene. 
           The camera follows a convoluted path on street level and encounters many turns, crossings, other cars and varying lighting conditions.
          }
  \label{fig:fakekitti_birdseye}%
\end{figure}%

\section{Introduction}%

Due to space limitations in the paper, this supplemental material contains a more detailed description of the dataset generation process (Section~\ref{sec:datageneration}) as well as more details and more qualitative results of DispNet (Section~\ref{sec:dispnet}).

\section{Dataset creation details}\label{sec:datageneration}%

\begin{figure}[t]
  \begin{center}
  {%
    \setlength{\tabcolsep}{1pt}%
    \begin{tabular}{cccc}%
      & Frame $t\!-\!1$ & Frame $t$ & Frame $t\!+\!1$ \\
      
      &
      \includegraphics[width=0.305\linewidth]{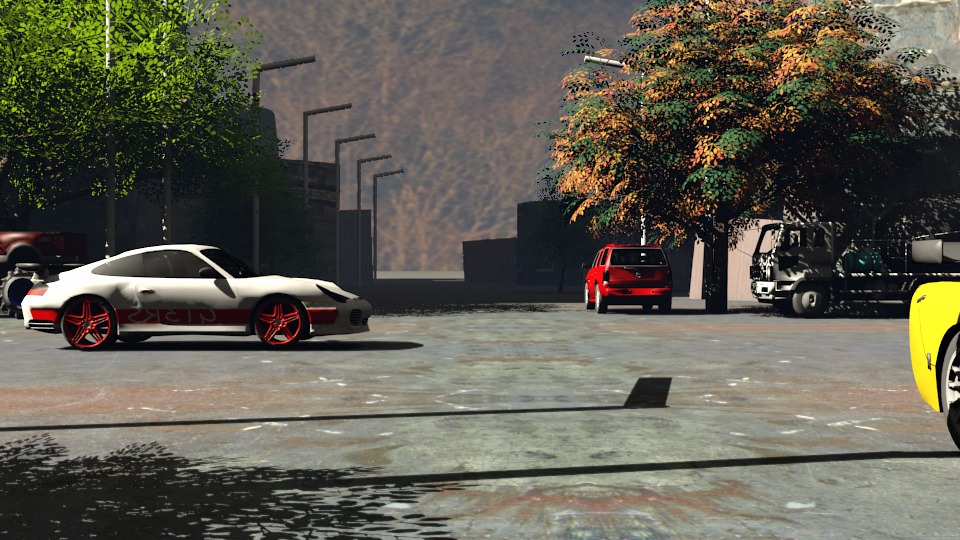} &
      {%
        \setlength{\fboxsep}{0pt}%
        \setlength{\fboxrule}{2pt}%
        \fbox{\includegraphics[width=0.305\linewidth]{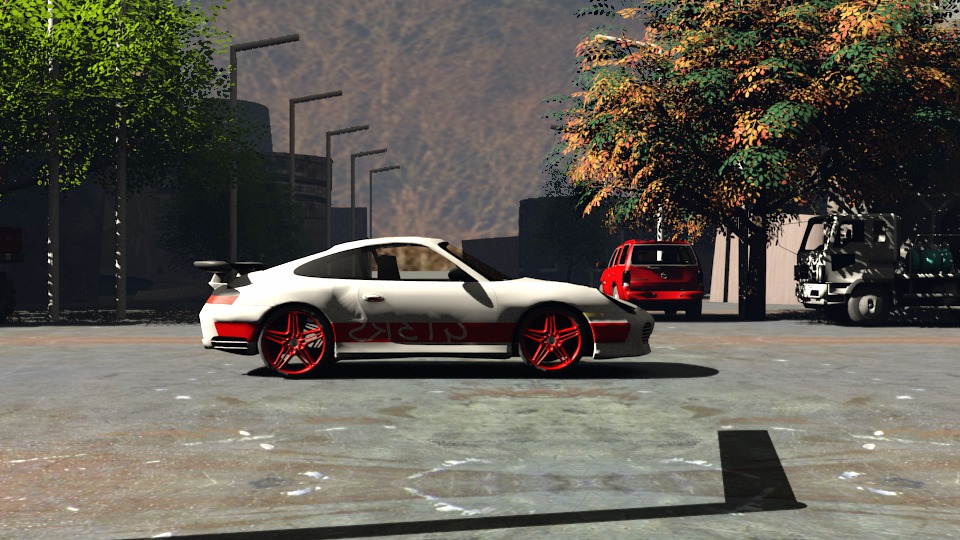}}%
      }&
      \includegraphics[width=0.305\linewidth]{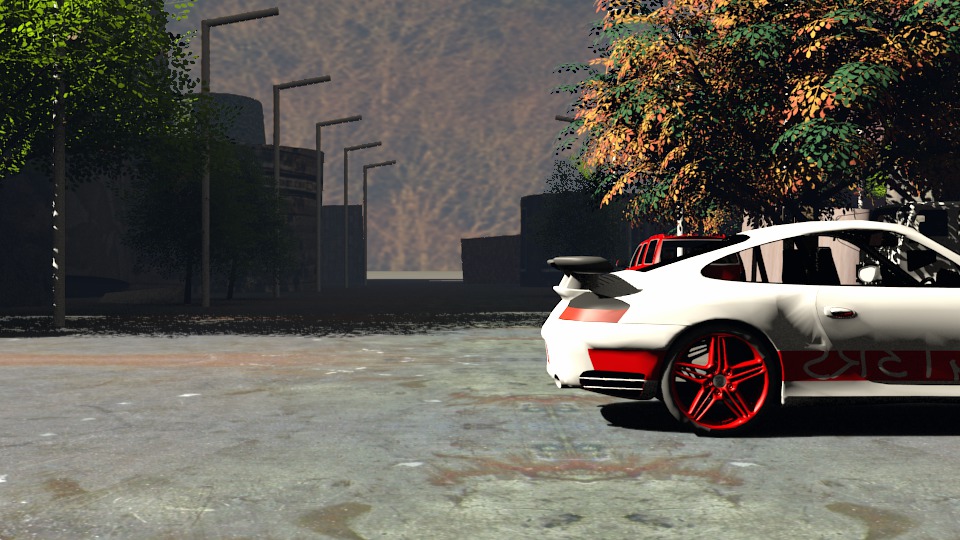} \\
      
      &
      \multicolumn{3}{c}{%
        \raisebox{-10pt}{ \!\! $\overbrace{\phantom{a+a+a+a+a+a+a+a+a+a+a+a+a}}$ }%
      } \\
      
      & 3DPos$_{t-1}$ & 3DPos$_t$ & 3DPos$_{t+1}$ \\
      
      \raisebox{15pt}{$X$} &
      \includegraphics[width=0.305\linewidth]{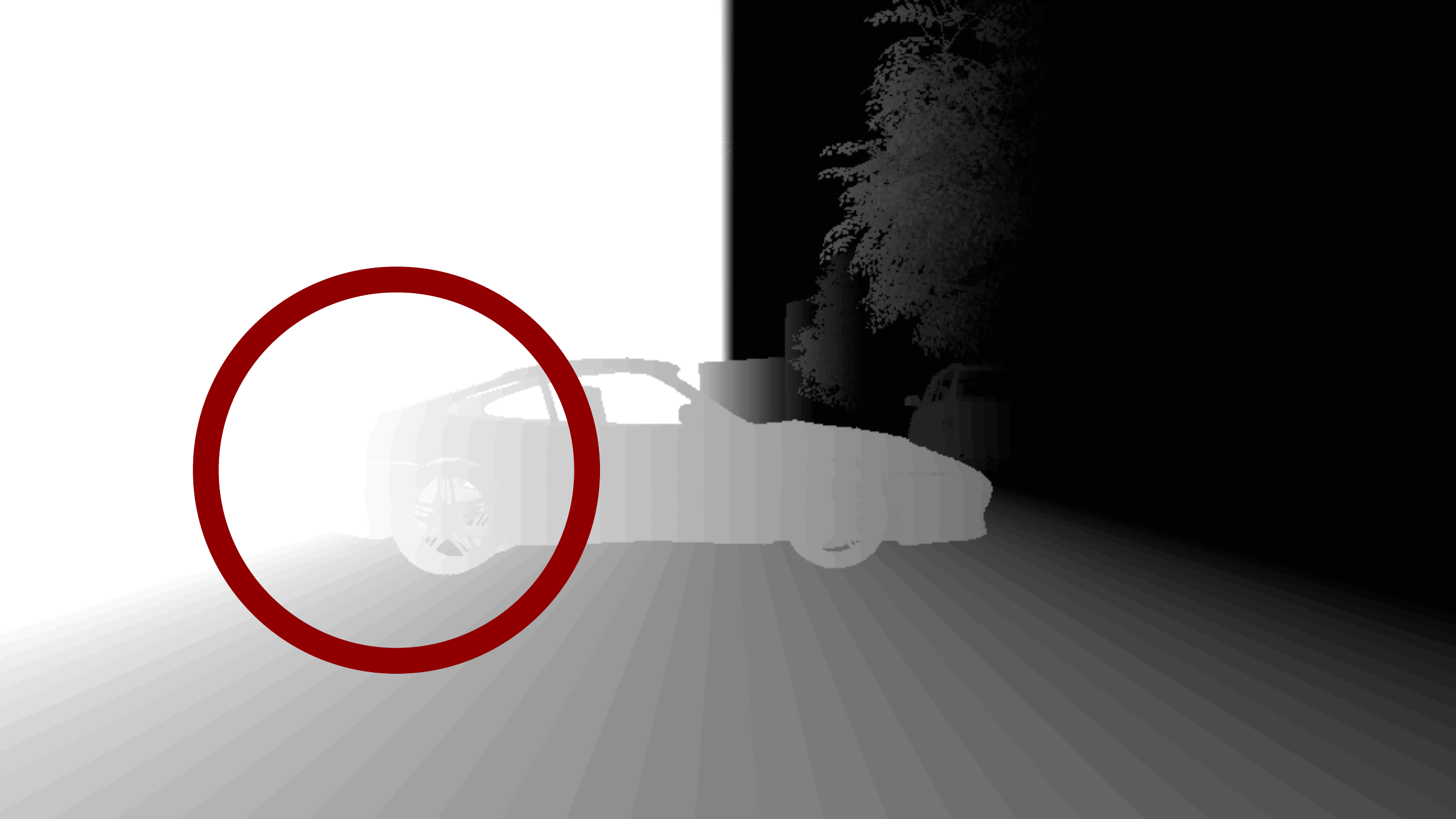} &
      \includegraphics[width=0.305\linewidth]{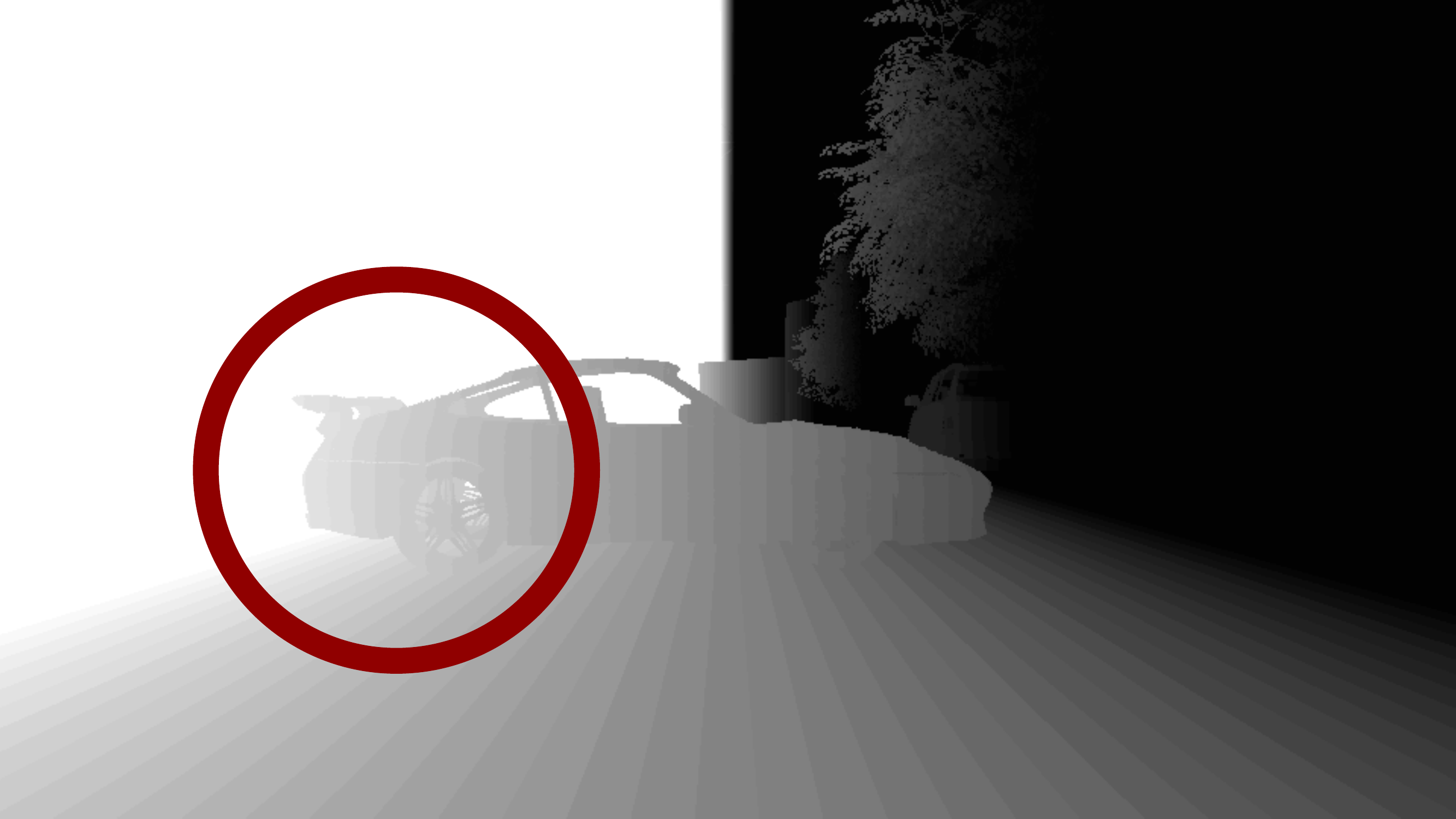} &
      \includegraphics[width=0.305\linewidth]{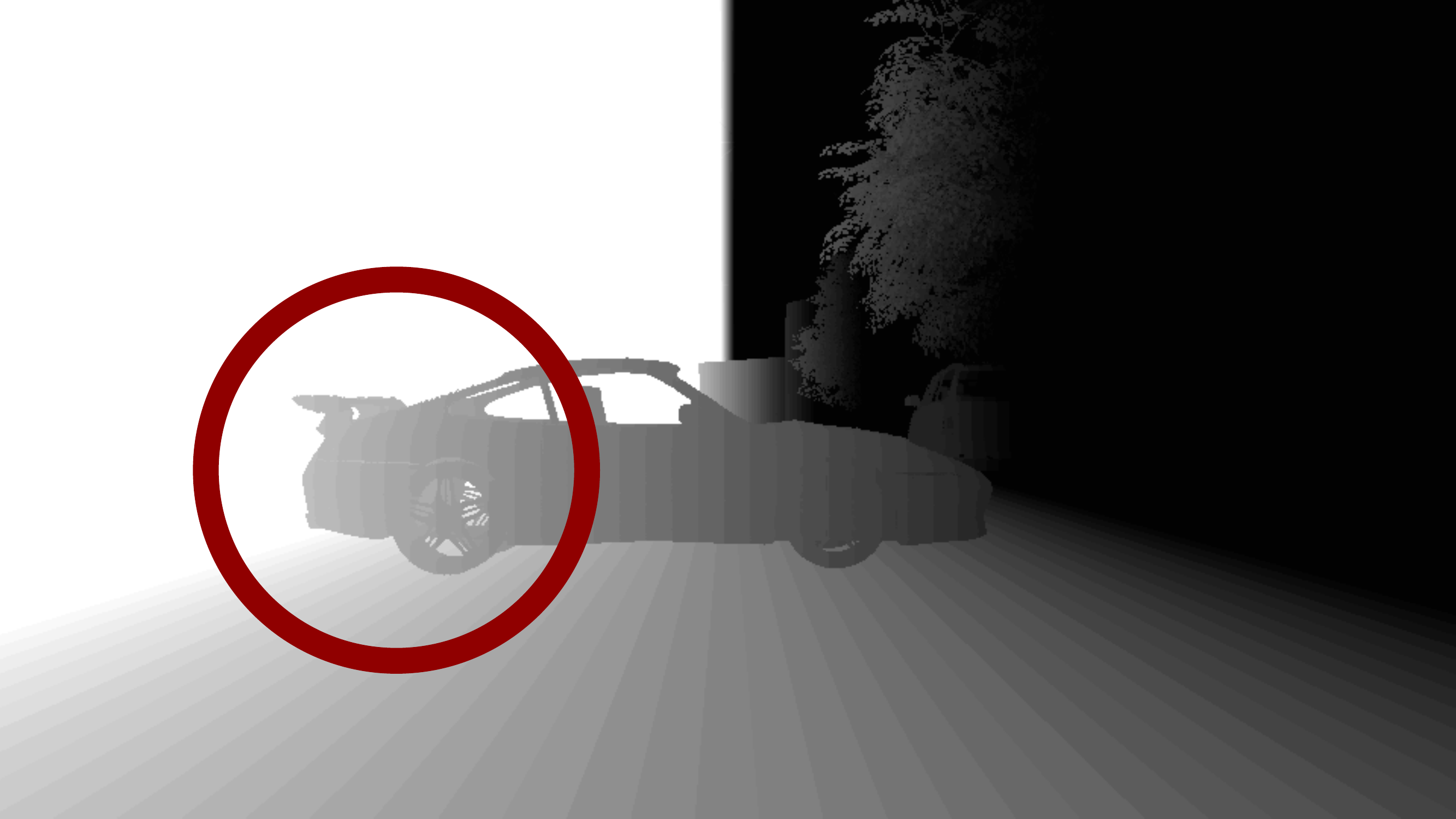} \\
      
      \raisebox{15pt}{$Y$} &
      \includegraphics[width=0.305\linewidth]{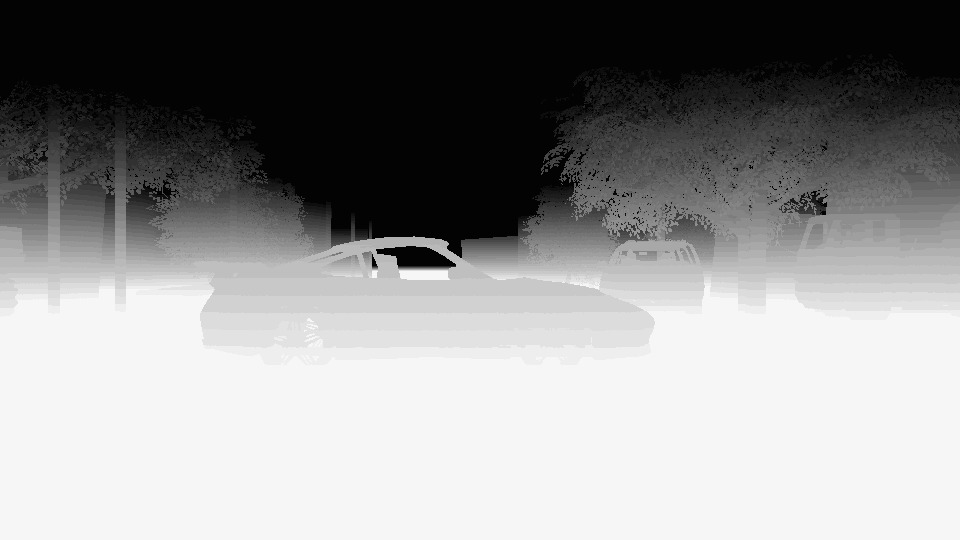} &
      \includegraphics[width=0.305\linewidth]{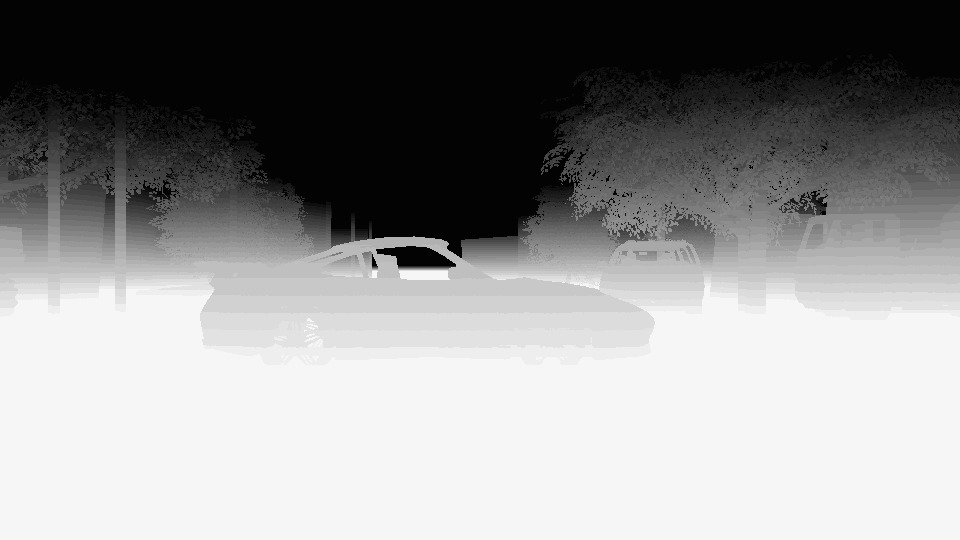} &
      \includegraphics[width=0.305\linewidth]{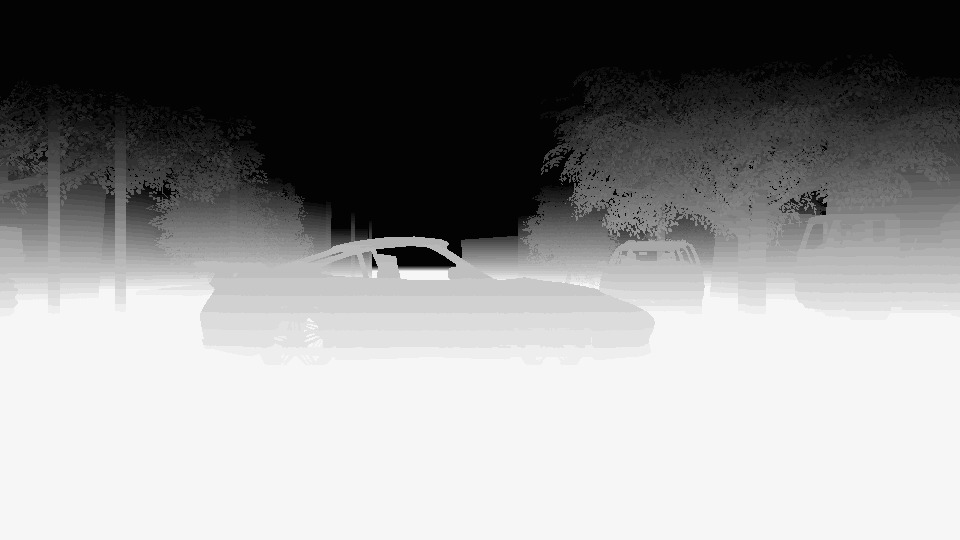} \\
      
      \raisebox{15pt}{$Z$} &
      \includegraphics[width=0.305\linewidth]{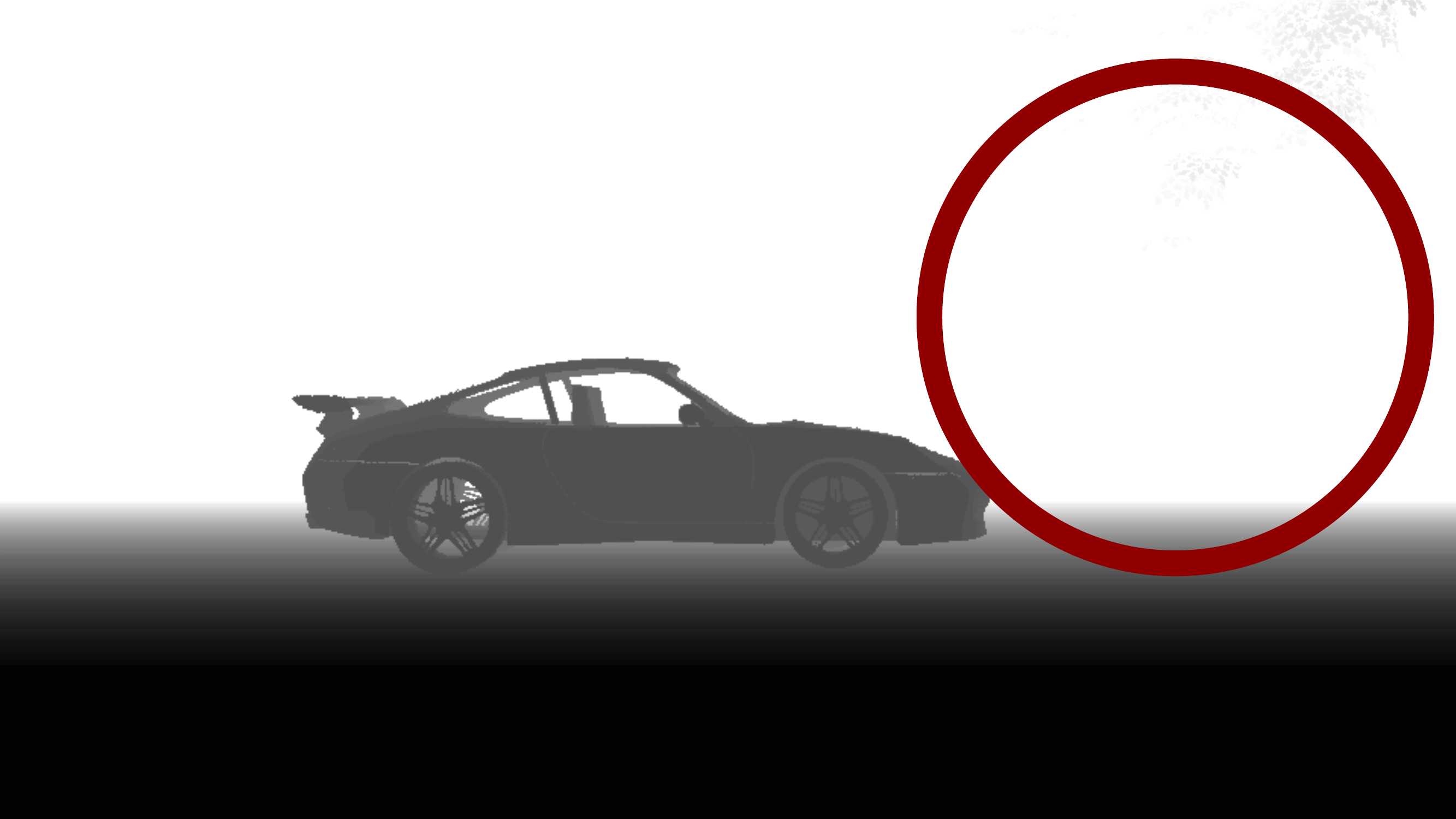} &
      \includegraphics[width=0.305\linewidth]{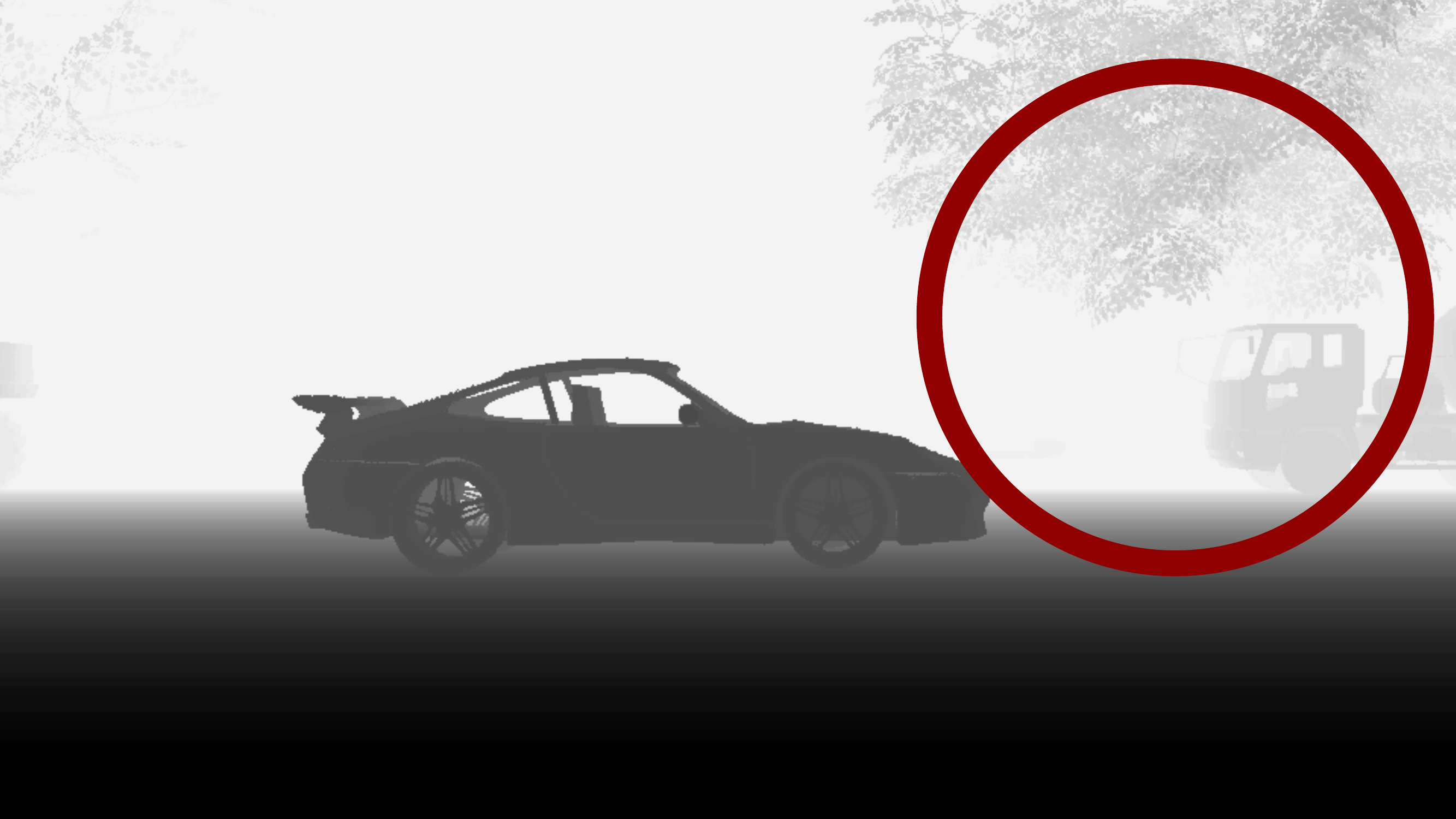} &
      \includegraphics[width=0.305\linewidth]{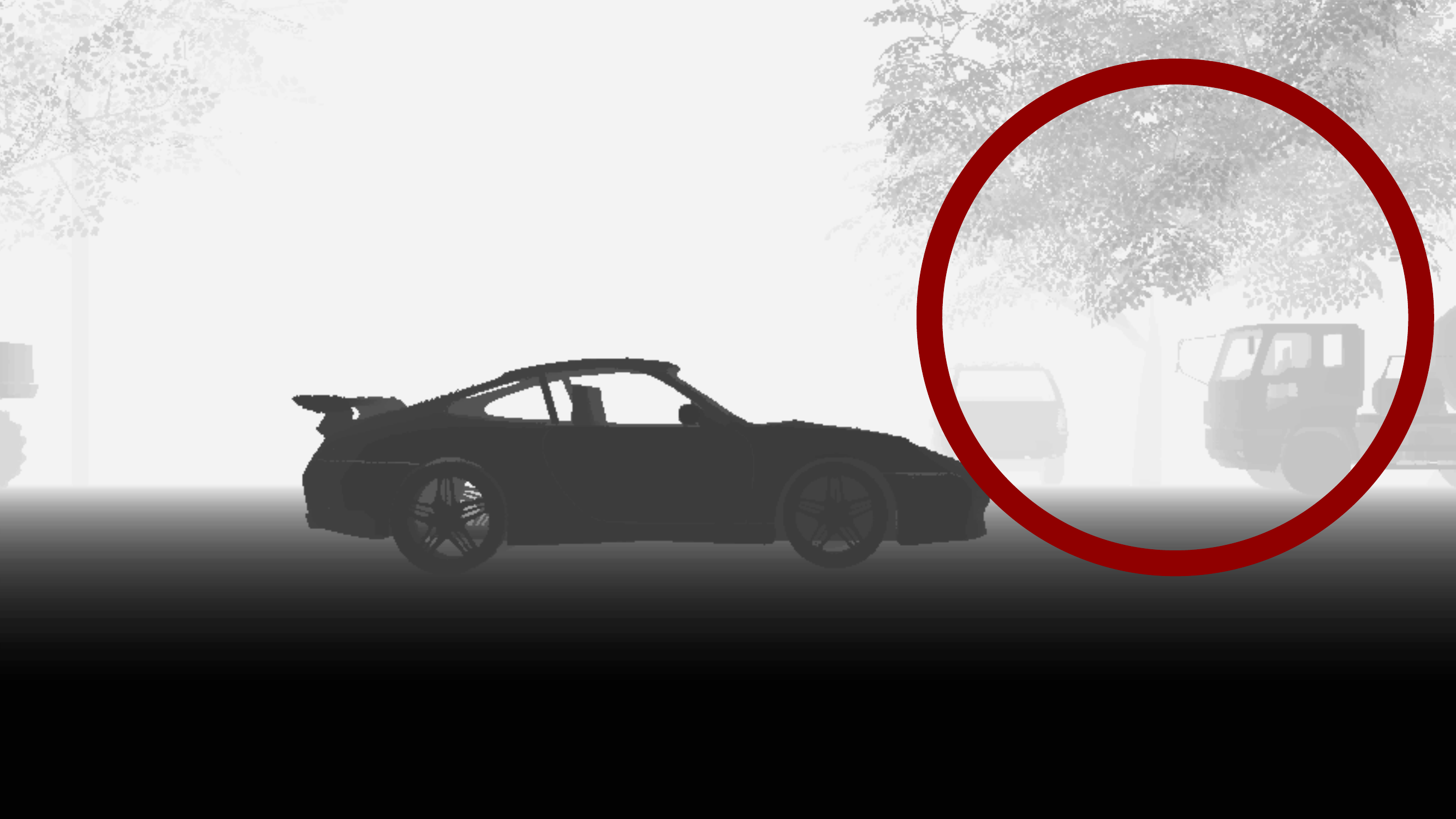} \\
      
      & Past & Now & Future \\
      
    \end{tabular}%
  }%
  \end{center}%
  \caption{Our intermediate render data for frame $t$:
           The \emph{X/Y/Z} channels encode the 3D positions (relative to the camera) of all visible points at frame $t$ (center column) and what their respective 3D positions were/will be in the previous/next frame (left/right columns).
           The 3D positions of the previous and next frame are stored at the same image locations as in frame $t$. Hence, analyzing a location from frame $t$ gives information about the past, current and future 3D position of the corresponding 3D point. All scene flow data can then be derived from this information.
           E.g.: The car moving to the right changes its $X$ values (note, that the perspective projection compresses the intensity gradient of the distant sky into an apparent step at $X=0$).
           Nothing is moving vertically, so all $Y$ values are constant over time.
           The camera is moving forward and all $Z$ values change uniformly (note, how objects on the right side become visible). 
          }%
  \label{fig:pos3d-images}
\end{figure}

We modified the pipeline of Blender's\footnote{Our modifications branch off of version 2.75b of the Blender source code. Starting with version 2.75, Blender supports multiview rendering.} internal render engine to produce -- besides stereo RGB images -- three additional data passes per frame and stereo view.
Fig.~\ref{fig:pos3d-images} gives a visual breakdown of this data:
\begin{itemize}%
  \item In the base pass (3DPos$_t$), each pixel stores the true 3D position of the scene point which projects into that pixel (the 3D position is given within the camera coordinate system).

  \item For the second pass (3DPos$_{t-1}$), we revert time to the previous frame~$t\!-\!1$ and save all vertices' 3D positions at that time.
    We then return to the current frame~$t$ and use the vertex 3D positions at time~$t$ to project the 3D vertices of time~$t-1$ into image space. 
    Hence, we again store 3D positions for each pixel, but this time \emph{the 3D positions from time $t\!-\!1$ using the projection at time~$t$}.
  \item The third pass (3DPos$_{t+1}$) is analogous to the second pass, except that this time we use the subsequent frame~$t\!+\!1$ instead of the previous frame~$t\!-\!1$.
\end{itemize}%

\begin{figure*}%
  \begin{center}%
    \resizebox{0.98\linewidth}{!}{%
      \includegraphics[]{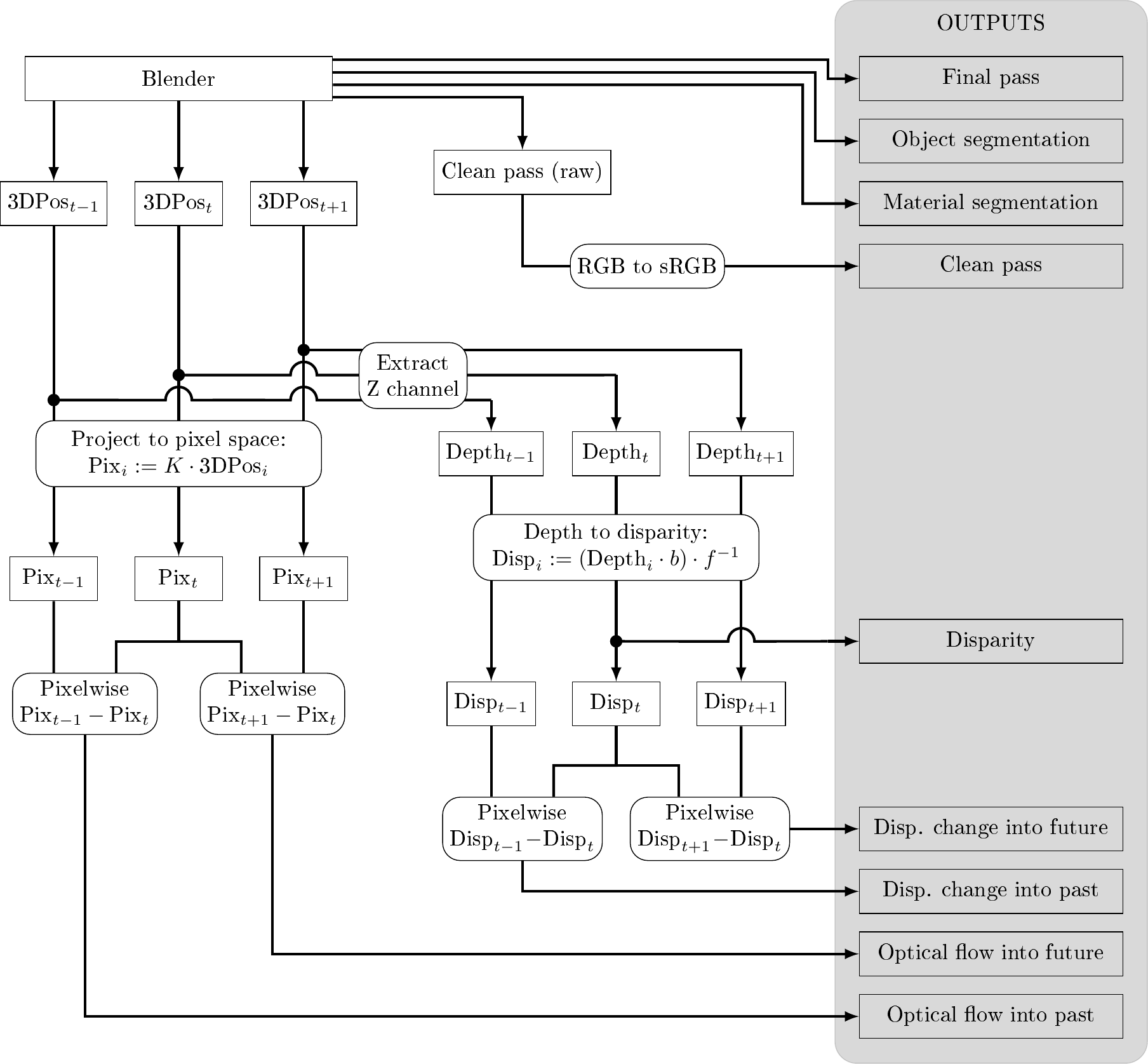}%
    }%
  \end{center}%
  \caption{\textbf{Data generation overview for a single view at frame time $\mathbf{t}$}: 
           Blender directly outputs the \emph{Final pass} and \emph{Clean pass} images, as well as the object-level and material-level \emph{segmentation masks}.
           \emph{Disparity} is directly obtained from depth, which is given by the $Z$ channel of the current 3DPos map as described in Fig.~\ref{fig:pos3d-images} ($b$ is the stereo baseline, $f$ denotes the focal length).
           Subtracting the current disparity map from the future/past disparity map results in the \emph{disparity change} in future/past direction.
           The original 3DPos images are projected from camera space into pixel space using the camera intrinsics matrix $K$.
           Subtracting the current pixel position image from the future/past pixel position images yields the \emph{optical flow} into the future/past.
          }%
  \label{fig:data_generation_flowchart}%
\end{figure*}

These three data structures contain all information about the 3D structure and 3D motion of the scene as seen from the current viewpoint. 
From the 3DPos data we generate the scene flow data.
Fig.~\ref{fig:data_generation_flowchart} describes the data conversion steps from the blender output to the resulting dataset. 
Note that color images and segmentation masks are directly produced by Blender and do not need any post-processing. 
Together with the camera intrinsics and extrinsics, various data can be generated, including calibrated RGBD images.


\begin{figure}[t]
  \begin{center}
  {
    \setlength{\tabcolsep}{1pt}%
    \begin{tabular}{ccc}
      \includegraphics[width=0.325\linewidth]{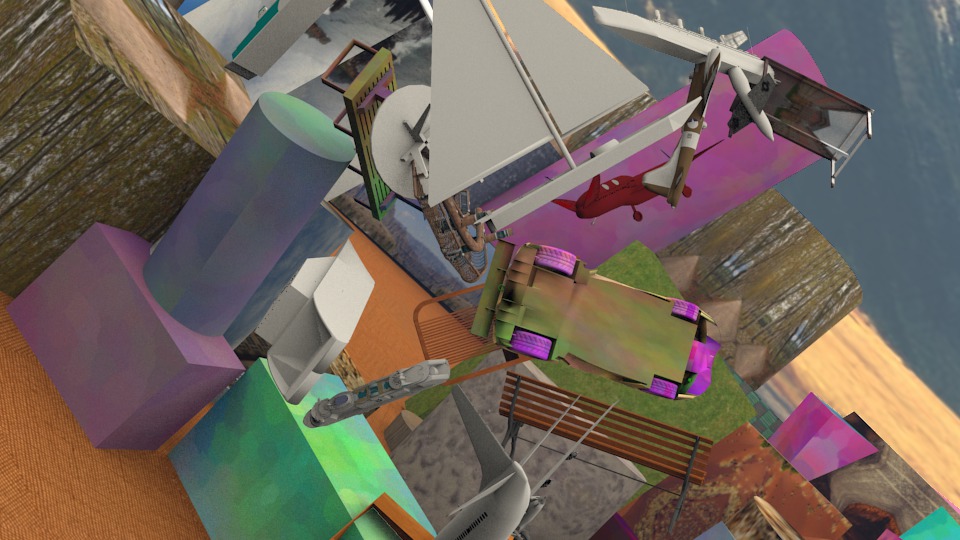} &
      \includegraphics[width=0.325\linewidth]{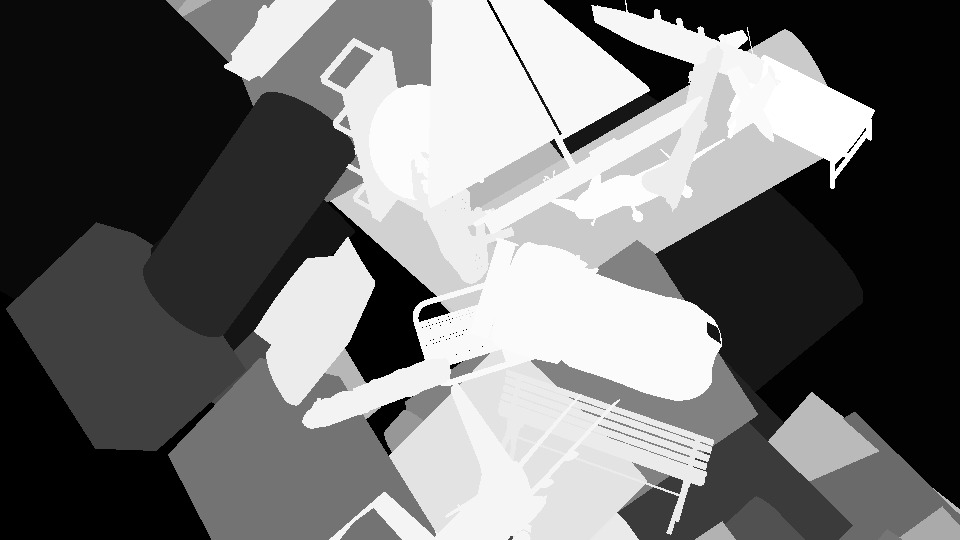} &
      \includegraphics[width=0.325\linewidth]{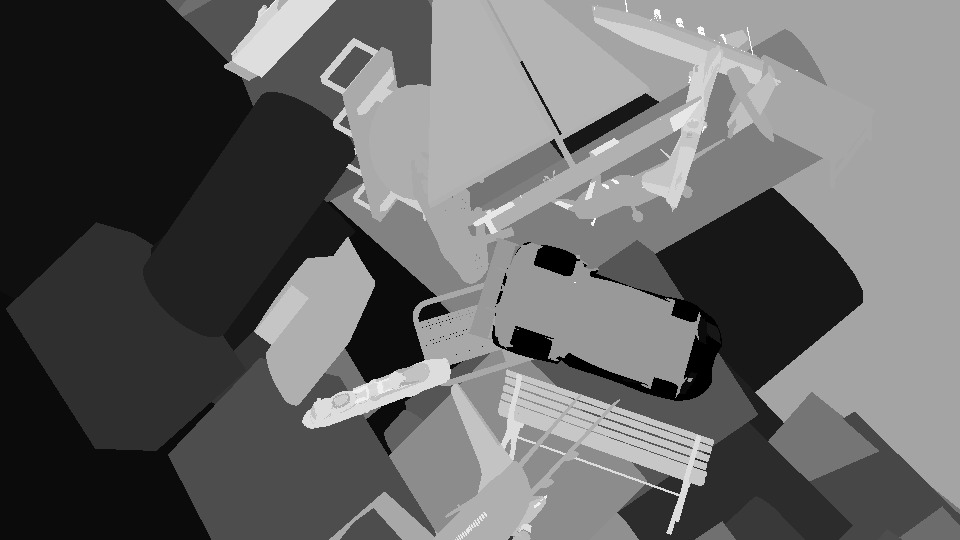} \\
      Color image & Object indices & Material indices \\
    \end{tabular}
  }
  \end{center}
  \caption{Segmentation data: 
          object indices are unique per scene. 
          Material indices can be shared across objects, but can be combined with the object indices to yield an oversegmentation into parts.
          }
  \label{fig:index-images}
\end{figure}

Fig.~\ref{fig:index-images} shows example segmentation masks for a frame from one of our datasets.
Materials can be shared across objects, but the combination of object indices and material indices yields a unique oversegmentation of a scene (consistent across all frames of the scene).
While our experiments do not make use of these data, for other applications we also include the object and material IDs in our dataset.

With this supplemental material, we also provide a video that demonstrates the
datasets we created and the final outcome of the pipeline, i.e. optical flow, disparity, disparity change and object and material index ground truth. 

\section{DispNetCorr}\label{sec:dispnet}%

Intuitively, the simple \emph{DispNet} disparity estimation architecture (as described in the main paper) has to learn the concept of matching parts of different images in rectified stereo images from scratch.
Since the structure of the problem is well known (correspondences can only be found in accordance with the epipolar geometry~\cite{multipleviewgeometry}), we introduced an alternative architecture -- the \emph{DispNetCorr} -- in which we explicitly correlate features along horizontal scanlines.

While the \emph{DispNet} uses two stacked RGB images as a single input (i.e. one six-channel input blob),
the \emph{DispNetCorr} architecture first processes the input images separately, then correlates features between the two images and further processes the result.
This behavior is similar to the correlation architecture used in~\cite{FlowNet} where Dosovitskiy~et~al. constructed a 2D correlation layer with limited neighborhood size and different striding in each of the images.
For disparity estimation, we can use a simpler approach without striding and with larger neighborhood size, because the correlation along one dimension is computationally less demanding. 
One can additionally reduce the amount of comparisons by limiting the search to only one direction.
For example, if we are given a left camera image and look for correspondences within the right camera image, then all disparity displacements are to the left.

Given two feature blobs $\mathbf{a}$ and $\mathbf{b}$ with multiple channels and identical sizes, we compute a correlation map of the same width and height, but with $D$ channels, where $D$ is the number of possible disparity values. For one pixel at location $(x,y)$ in the first feature blob $\mathbf{a}$, the resulting correlation entry at channel $d\!\in\![0,D-1]$ is the scalar product of the two feature vectors $\mathbf{a}_{(x,y)}$ and $\mathbf{b}_{(x-d,y)}$.

\section{Qualitative Examples}\label{sec:qual}
We show a qualitative evaluation of our networks for disparity estimation and compare them to other approaches in Figures~\ref{fig:gallery_sintel_0} to~\ref{fig:gallery_kitti_131}.

\newcommand{\galleryfigure}[2]{
\begin{figure*}[h]
  \begin{center}%
  \resizebox{0.99\linewidth}{!}{%
    \setlength{\tabcolsep}{1pt}%
    \begin{tabular}{ccccc}%
      RGB image (L) &
      RGB image (R) &
      DispNet &
      DispNet-K &
      MC-CNN-fst \\
      \includegraphics[width=0.245\linewidth]{gallery/#1_imgL.jpg} &
      \includegraphics[width=0.245\linewidth]{gallery/#1_imgR.jpg} &
      \includegraphics[width=0.245\linewidth]{gallery/#1_A.jpg} &
      \includegraphics[width=0.245\linewidth]{gallery/#1_C.jpg} &
      \includegraphics[width=0.245\linewidth]{gallery/#1_E.jpg} \\

      \multicolumn{2}{l}{\includegraphics[width=0.245\linewidth]{gallery/#1_gt.jpg}} &
      \includegraphics[width=0.245\linewidth]{gallery/#1_B.jpg} &
      \includegraphics[width=0.245\linewidth]{gallery/#1_D.jpg} &
      \includegraphics[width=0.245\linewidth]{gallery/#1_F.jpg} \\
      ground truth (sparse)&
      &
      DispNetCorr1D &
      DispNetCorr1D-K &
      SGM \\
    \end{tabular}%
  }%
  \end{center}%
  \caption{#2}%
  \label{fig:gallery_#1}%
\end{figure*}
}

\newcommand{\galleryfigureThreeByThree}[2]{
\begin{figure*}[h]
  \begin{center}%
  \resizebox{0.99\linewidth}{!}{%
    \setlength{\tabcolsep}{1pt}%
    \begin{tabular}{ccc}%
      RGB image (L) &
      DispNet &
      DispNetCorr1D \\
      \includegraphics[width=0.33\linewidth]{gallery/#1_imgL.jpg} &
      \includegraphics[width=0.33\linewidth]{gallery/#1_A.jpg} &
      \includegraphics[width=0.33\linewidth]{gallery/#1_B.jpg} \\
      RGB image (R) &
      DispNet-K &
      DispNetCorr1D-K \\
      \includegraphics[width=0.33\linewidth]{gallery/#1_imgR.jpg} &
      \includegraphics[width=0.33\linewidth]{gallery/#1_C.jpg} &
      \includegraphics[width=0.33\linewidth]{gallery/#1_D.jpg} \\
      ground truth &
      MC-CNN-fst &
      SGM \\
      \includegraphics[width=0.33\linewidth]{gallery/#1_gt.jpg} &
      \includegraphics[width=0.33\linewidth]{gallery/#1_E.jpg} &
      \includegraphics[width=0.33\linewidth]{gallery/#1_F.jpg} \\
    \end{tabular}%
  }
  \end{center}%
  \caption{#2}%
  \label{fig:gallery_#1}%
\end{figure*}
}

\newcommand{\galleryfigureWithOcc}[2]{
\begin{figure*}[h]
  \begin{center}%
  \resizebox{0.99\linewidth}{!}{%
    \setlength{\tabcolsep}{1pt}%
    \begin{tabular}{ccccc}%
      RGB image (L) &
      RGB image (R) &
      DispNet &
      DispNet-K &
      MC-CNN-fst \\
      \includegraphics[width=0.245\linewidth]{gallery/#1_imgL.jpg} &
      \includegraphics[width=0.245\linewidth]{gallery/#1_imgR.jpg} &
      \includegraphics[width=0.245\linewidth]{gallery/#1_A.jpg} &
      \includegraphics[width=0.245\linewidth]{gallery/#1_C.jpg} &
      \includegraphics[width=0.245\linewidth]{gallery/#1_E.jpg} \\

      \includegraphics[width=0.245\linewidth]{gallery/#1_gt.jpg} &
      \includegraphics[width=0.245\linewidth]{gallery/#1_occ_gt.jpg} &
      \includegraphics[width=0.245\linewidth]{gallery/#1_B.jpg} &
      \includegraphics[width=0.245\linewidth]{gallery/#1_D.jpg} &
      \includegraphics[width=0.245\linewidth]{gallery/#1_F.jpg} \\
      ground truth &
      occlusion ground truth&
      DispNetCorr1D &
      DispNetCorr1D-K &
      SGM \\
    \end{tabular}%
  }%
  \end{center}%
  \caption{#2}%
  \label{fig:gallery_#1}%
\end{figure*}
}

\galleryfigureWithOcc{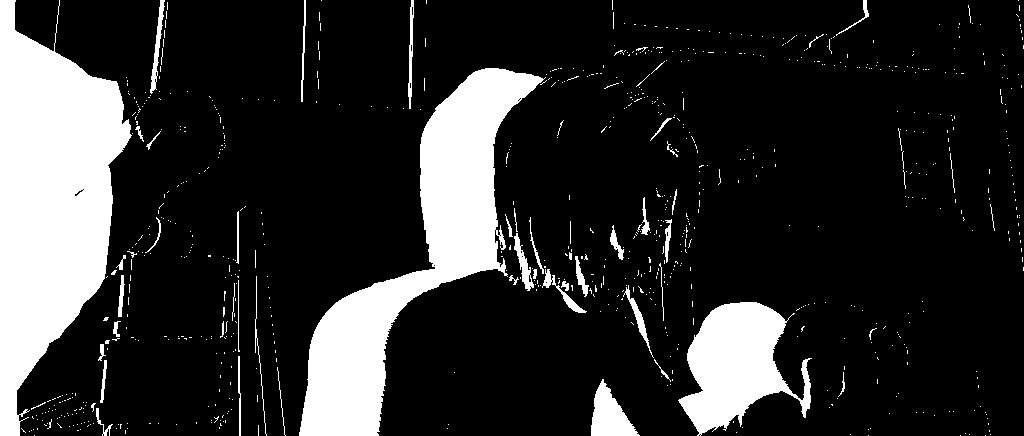}{
Disparities on a Sintel frame:
DispNet and DispNetCorr1D fill the occluded regions in a much more reasonable way compared to other approaches.}

\galleryfigureWithOcc{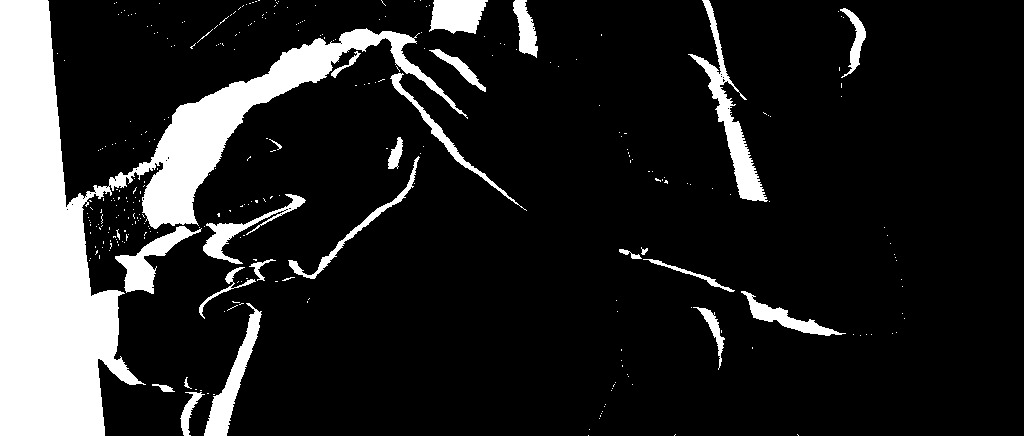}{
Disparities on a Sintel frame: 
DispNetCorr1D provides sharper estimates and the smooth areas on the dragon head are estimated better than with DispNet.}

\galleryfigureWithOcc{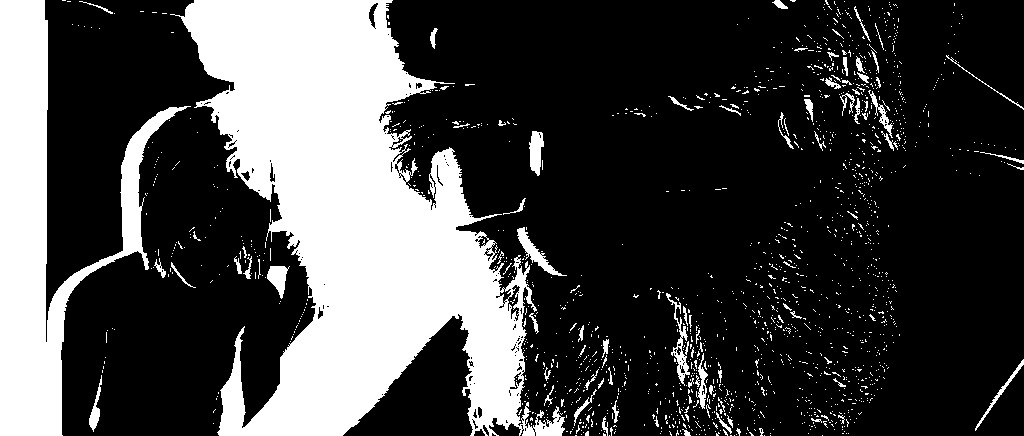}{
Disparities on a Sintel frame: 
The networks finetuned on the KITTI 2015 dataset cannot estimate large disparities anymore (large disparities are not present in KITTI). Also MC-CNN-fst has 
problems with the large disparities.}

\galleryfigureWithOcc{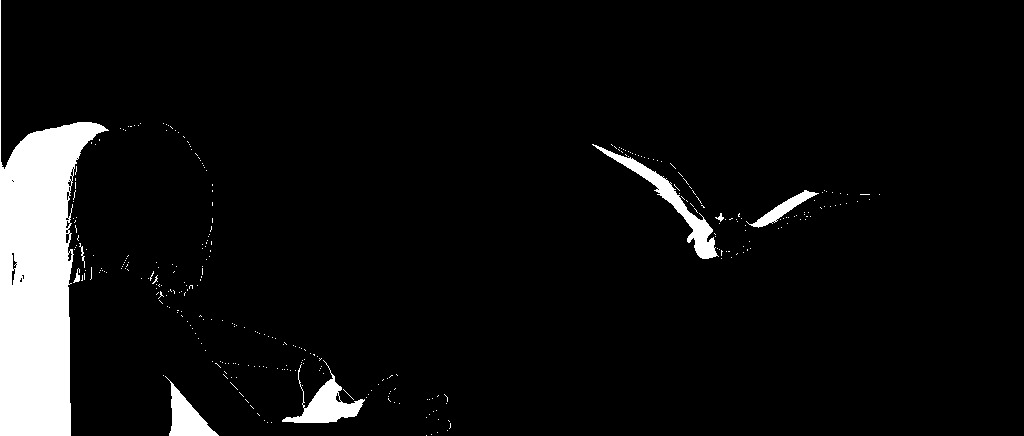}{
Disparities on a Sintel frame: 
DispNet and DispNetCorr1D can handle occluded regions in a nice way.
After finetuning on KITTI 2015 the networks fail in the sky region (ground truth for sky and other small disparities are not present in KITTI).}

\galleryfigureThreeByThree{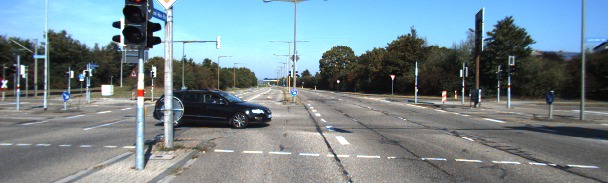}{
Disparities on a KITTI 2015 frame: 
The sparsity of the KITTI 2015 dataset leads to very smooth predictions when finetuning a network with such ground truth. While the non-finetuned DispNet and DispNetCorr1D estimate fine details accurately, they are less accurate in the smooth road and ground regions which are very common in KITTI.}

\galleryfigureThreeByThree{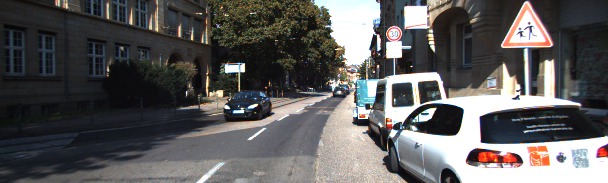}{
Disparities on a KITTI 2015 frame: 
Finetuning the networks on KITTI leads to much smoother estimates. However, DispNet-K and DispNetCorr1D-K can still recognize the delineator posts in the bottom left, which DispNet and DispNetCorr1D ignore completely. This shows that the finetuned networks do not simply oversmooth, but are still able to find small structures and disparity discontinuities. }

\end{document}